\documentclass[10pt,twocolumn,letterpaper]{article}

\usepackage{cvpr}
\usepackage{times}
\usepackage{epsfig}
\usepackage{graphicx}
\usepackage{amsmath}
\usepackage{amssymb}
\usepackage{subfig} 

\usepackage{color}
\usepackage{multirow}
\usepackage{url}
\usepackage{epstopdf}
\usepackage{ctable}
\usepackage{multirow}
\usepackage{flushend}
\usepackage{array}
\usepackage{appendix}

\usepackage[pagebackref=true,breaklinks=true,letterpaper=true,colorlinks,bookmarks=false]{hyperref}

\graphicspath{{figures/}}

\cvprfinalcopy 

\def\cvprPaperID{3134} 

\newcommand{\specialcell}[2][c]{%
	  \begin{tabular}[#1]{@{}c@{}}#2\end{tabular}}

\ifcvprfinal\fi
\begin{document}

\title{Learning Diverse Image Colorization}

\author{
Aditya Deshpande, Jiajun Lu, Mao-Chuang Yeh, Min Jin Chong and David Forsyth\\
University of Illinois at Urbana Champaign\\
{\tt\small \{ardeshp2, jlu23, myeh2, mchong6, daf\} @illinois.edu}
}

\maketitle

\begin{abstract}
Colorization is an ambiguous problem, with multiple viable colorizations for a single grey-level
image. However, previous methods only produce the single most probable colorization. Our goal
is to model the diversity intrinsic to the problem of colorization and produce multiple colorizations
that display long-scale spatial co-ordination. We learn a low dimensional embedding of 
color fields using a variational autoencoder (VAE). We construct loss terms for the VAE decoder that avoid
blurry outputs and take into account the uneven distribution of pixel colors. Finally, we build
a conditional model for the multi-modal distribution between grey-level image and the
color field embeddings. Samples from this conditional model result in diverse colorization.
We demonstrate that our method obtains better diverse colorizations than a standard 
conditional variational autoencoder (CVAE) model, as well as a recently proposed conditional 
generative adversarial network (cGAN).
\end{abstract}


\section{Introduction}
\label{sec:intro}

In colorization, we predict the 2-channel color field for an input
grey-level image. It is an inherently ill-posed and an ambiguous 
problem. Multiple different colorizations are possible for a 
single grey-level image. For example, different shades of blue 
for sky, different colors for a building, different skin tones for a person 
and other stark or subtle color changes  are all acceptable colorizations. 
In this paper, our goal is to generate multiple colorizations for a single grey-level image that are 
diverse and at the same time, each realistic. This is a 
demanding task, because color fields are not only cued to the
local appearance but also have a long-scale spatial structure.
Sampling colors independently from per-pixel distributions 
makes the output spatially incoherent and it does not generate a 
realistic color field (See Figure \ref{fig:rand_sample}). Therefore, we need a method that generates
multiple colorizations while balancing per-pixel color 
estimates and long-scale spatial co-ordination. This paradigm
is common to many ambiguous vision tasks where multiple 
predictions are desired viz. generating motion-fields from
static image \cite{Walker}, synthesizing future frames \cite{CrossConv}, 
time-lapse videos \cite{TLBerg}, interactive segmentation and
pose-estimation \cite{Batra} etc.

A natural approach to solve the problem is to learn a conditional 
model $P(\mathbf{C}|\mathbf{G})$ for a color field $\mathbf{C}$ conditioned 
on the input grey-level image $\mathbf{G}$. We can then draw samples from 
this conditional model $\{\mathbf{C}_k\}_{k=1}^{N} \sim P(\mathbf{C}|\mathbf{G})$ 
to obtain diverse colorizations. To build this explicit conditional model 
is difficult. The difficulty being $\mathbf{C}$ and $\mathbf{G}$ are 
high-dimensional spaces. The distribution of natural color fields and 
grey-level features  in these high-dimensional spaces is therefore scattered. This does not
expose the sharing required to learn a multi-modal conditional model. Therefore, 
we seek feature representations of $\mathbf{C}$ and $\mathbf{G}$ that allow us to 
build a conditional model.

\begin{figure*}[t]
\centering
\includegraphics[trim={0 5.2cm 0 0},clip,width=.95\textwidth]{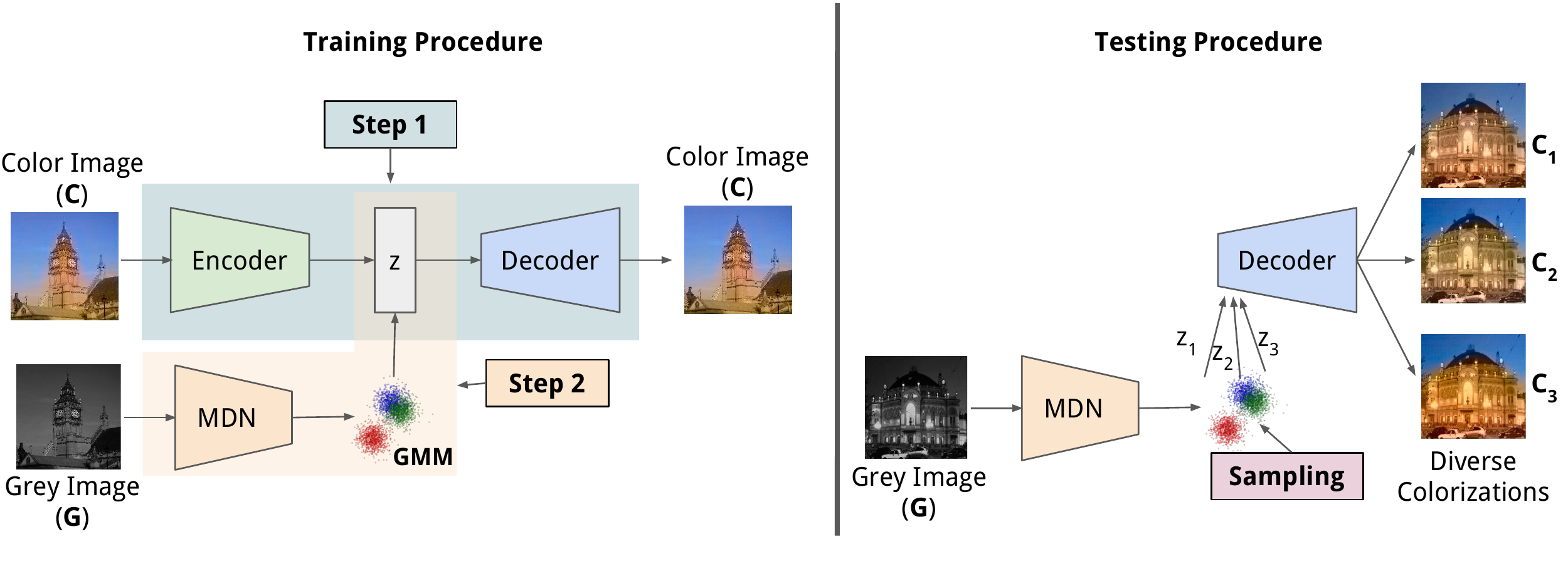}
\caption{Step 1, we learn a low-dimensional embedding $\mathbf{z}$ for a color field 
$\mathbf{C}$. Step 2, we train a multi-modal conditional model $P(\mathbf{z}|\mathbf{G})$ 
that generates the low-dimensional embedding from grey-level features $\mathbf{G}$. At test time,
we can sample the conditional model $\{ \mathbf{z}_{k} \}_{k=1}^{N} \sim P(\mathbf{z}|\mathbf{G})$ 
and use the VAE decoder to generate the corresponding diverse color fields $\{ \mathbf{C}_k \}_{k=1}^{N}$.} 
\label{fig:overview} 
\end{figure*}

Our strategy is to represent $\mathbf{C}$ by its low-dimensional
latent variable embedding $\mathbf{z}$. This embedding is learned 
by a generative model such as the Variational Autoencoder (VAE) 
\cite{AEB} (See Step 1 of Figure \ref{fig:overview}). Next, we leverage 
a Mixture Density Network (MDN) to learn a multi-modal conditional 
model $P(\mathbf{z}|\mathbf{G})$ (See Step 2 of Figure \ref{fig:overview}). 
Our feature representation for grey-level image $\mathbf{G}$ comprises 
the features from conv-7 layer of a colorization CNN \cite{ZhangColorful}. 
These features encode spatial structure and per-pixel affinity to colors. Finally, 
at test time we sample multiple $\{ \mathbf{z}_{k} \}_{k=1}^{N} \sim P(\mathbf{z}|\mathbf{G})$ 
and use the VAE decoder to obtain the corresponding colorizations $\mathbf{C}_k$
for each $\mathbf{z}_k$ (See Figure \ref{fig:overview}). Note that, 
our low-dimensional embedding encodes the spatial structure of color fields 
and we obtain spatially coherent diverse colorizations by sampling the 
conditional model.

The contributions of our work are as follows. First, we learn 
a smooth low-dimensional embedding along with a device to 
generate corresponding color fields with high fidelity 
(Section \ref{sec:vae}, \ref{sec:res_vae_loss}). Second, we a learn multi-modal conditional 
model between the grey-level features and the low-dimensional 
embedding capable of producing diverse colorizations (Section \ref{sec:mdn}). Third, we show that our
method outperforms the strong baseline of conditional 
variational autoencoders (CVAE) and conditional generative adversarial 
networks (cGAN) \cite{Isola} for obtaining diverse colorizations 
(Section \ref{sec:res_div}, Figure \ref{fig:res_cvae_mdn_div}).

\section{Background and Related Work}

\begin{figure}[t]
\centering
\subfloat[Sampling per-pixel distribution of \cite{ZhangColorful}]{\includegraphics[width=.32\textwidth]{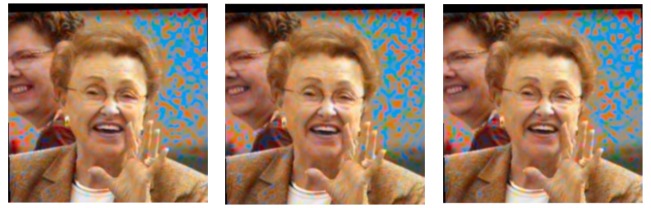}} \hspace{5pt} 
\subfloat[Ground truth]{\includegraphics[width=.105\textwidth]{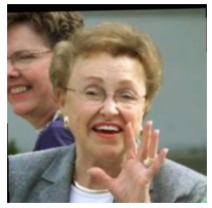}}
\caption{Zhang et al.\ \cite{ZhangColorful} predict a per-pixel probability
distribution over colors. First three images are diverse colorizations obtained 
by sampling the per-pixel distributions independently. The last image is the 
ground-truth color image. These images demonstrate the speckled noise and lack 
of spatial co-ordination resulting from independent sampling of pixel colors.} 
\label{fig:rand_sample} 
\end{figure}

\noindent
{\bf Colorization.} Early colorization methods were interactive,
they used a reference color image \cite{Welsh} or scribble-based
color annotations \cite{Levin}. Subsequently, \cite{Charpiat08ECCV,DeepCol,LSCol,Jancsary,Morimoto} performed
automatic image colorization without any human annotation or
interaction. However, these methods were trained on datasets
of limited sizes, ranging from a few tens to a few thousands 
of images. Recent CNN-based methods have been able to
scale to much larger datasets of a million images 
\cite{LetColor,Gustav16,ZhangColorful}. All these methods
are aimed at producing only a single color image as output.
\cite{Charpiat08ECCV,Gustav16,ZhangColorful} predict a multi-modal 
distribution of colors over each pixel. But, \cite{Charpiat08ECCV}
performs a graph-cut inference to produce a single color field
prediction, \cite{ZhangColorful} take expectation after making 
the per-pixel distribution peaky and \cite{Gustav16} sample the 
mode or take the expectation at each pixel to generate single 
colorization. To obtain diverse colorizations from 
\cite{Gustav16,ZhangColorful}, colors have to be sampled independently for 
each pixel. This leads to speckle noise in the output color fields 
as shown in Figure \ref{fig:rand_sample}. Furthermore, one obtains little
diversity with this noise. Isola et al.~\cite{Isola} use conditional
GANs for the colorization task. Their focus is to generate single
colorization for a grey-level input. We produce diverse 
colorizations for a single input, which are all realistic. \\ 

\noindent
{\bf Variational Autoencoder.}
As discussed in Section \ref{sec:intro}, we wish to learn a low-dimensional embedding $\mathbf{z}$ of 
a color field $\mathbf{C}$. Kingma and Welling \cite{AEB} 
demonstrate that this can be achieved using a variational 
autoencoder comprising of an encoder network and a decoder 
network. They derive the following lower bound on log likelihood:

\begin{equation}
\label{eq:vae_lowerbound}
\begin{aligned}
\mathbb{E}_{z \sim Q} \lbrack \log P(\mathbf{C}|\mathbf{z}, \theta)\rbrack - 
\mathcal{KL}\lbrack Q(\mathbf{z} | \mathbf{C}, \theta) \| P(\mathbf{z}) \rbrack
\end{aligned}
\end{equation}

The lower bound is maximized by maximizing \autoref{eq:vae_lowerbound}
with respect to parameters $\theta$. They assume the posterior $P(\mathbf{C}|\mathbf{z}, \theta)$ 
is a Gaussian distribution $\mathcal{N}(\mathbf{C} | f(\mathbf{z}, \theta), \sigma^{2})$. 
Therefore, the first term of \autoref{eq:vae_lowerbound} reduces to a decoder network $f(\mathbf{z}, \theta)$
with an L$_2$ loss $\| \mathbf{C} - f(\mathbf{z}, \theta)\|_{2}$. Further, they assume 
the distribution $P(\mathbf{z})$ is a zero-mean unit-variance Gaussian distribution. Therefore, the 
encoder network $Q(\mathbf{z}|\mathbf{C}, \theta)$ is trained with a KL-divergence loss 
to the distribution $\mathcal{N}(0, I)$. Sampling, $z \sim Q$, is performed with the 
re-parameterization trick to enable backpropagation and the joint training of 
encoder and decoder. VAEs have been used to embed and decode Digits 
\cite{Draw,SemiSup,AEB}, Faces \cite{DCIGN,Attr2Img} and more recently CIFAR images 
\cite{Draw,Iavae}. However, they are known to produce blurry and over-smooth 
outputs. We carefully devise loss terms that discourage blurry, greyish outputs and 
incorporate specificity and colorfulness (Section \ref{sec:vae}). \\

\section{Embedding and Decoding a Color Field}
\label{sec:vae}

We use a VAE to obtain a low-dimensional embedding for a color field. In addition 
to this, we also require an efficient decoder that generates a realistic color field from a given
embedding. Here, we develop loss terms for VAE decoder that avoid the over-smooth and washed
out (or greyish) color fields obtained with the standard L$_2$ loss. 

\subsection{Decoder Loss}
\label{sec:vae_dec_loss}

\noindent
{\bf Specificity.} Top-k principal components, $\mathbf{P}_k$, are the directions of 
projections with maximum variance in the high dimensional space of color fields. 
Therefore, producing color fields that vary primarily along the top-k principal components 
provides reduction in L$_2$ loss at the expense of specificity in generated color fields. To 
disallow this, we project the generated color field $f(\mathbf{z}, \theta)$ and ground-truth 
color field $\mathbf{C}$ along top-k principal components. We use $k=20$ in our implementation. Next, we divide the 
difference between these projections along each principal component by the corresponding 
standard deviation $\sigma_{k}$ estimated from training set. This encourages changes along 
all principal components to be on an equal footing in our loss. The 
residue is divided by standard deviation of the $k^{th}$ (for our case $20^{th}$) component. 
Write specificity loss $\mathcal{L}_{mah}$ using the squared sum of these distances and residue,

\begin{gather*}
\medmuskip=1mu
\thinmuskip=1mu
\thickmuskip=1mu
\label{eq:vae_loss_pca}
\mathcal{L}_{mah} = \sum\limits_{k=1}^{20} \frac{\| \lbrack \mathbf{C} - f({\mathbf{z}, \theta})\rbrack^{T} 
\mathbf{P}_{k} \|^{2}_{2}}{\sigma^{2}_{k}} \; + \; 
\frac{\| \mathbf{C}_{res} - f_{res}(\mathbf{z}, \theta)\|^{2}_{2}}{\sigma^{2}_{20}} \\
\mathbf{C}_{res} = \mathbf{C} - \sum\limits_{k=1}^{20} \mathbf{C}^{T}\mathbf{P}_k \mathbf{P}_k \\ 
\mathbf{f}_{res}(\mathbf{z}, \theta) = \mathbf{f(\mathbf{z}, \theta)} - \sum\limits_{k=1}^{20} \mathbf{f(\mathbf{z}, \theta)}^{T}\mathbf{P}_k \mathbf{P}_k,  
\end{gather*}

The above loss is a combination of Mahalanobis distance \cite{mah30} between vectors
$\lbrack \mathbf{C}^{T}\mathbf{P}_1, \mathbf{C}^{T}\mathbf{P}_2, \cdots, \mathbf{C}^{T}\mathbf{P}_{20} \rbrack$
and $\lbrack f(\mathbf{z}, \theta)^{T}\mathbf{P}_1, f(\mathbf{z}, \theta)^{T}\mathbf{P}_2, \cdots, f(\mathbf{z}, \theta)^{T}\mathbf{P}_{20} \rbrack$ with a diagonal covariance matrix $\Sigma = \text{diag}(\sigma_{k})_{k=1 \text{ to } 20}$ and an additional residual term. \\

\noindent
{\bf Colorfulness.} The distribution of colors in images is highly imbalanced, with more 
greyish colors than others. This biases the generative model to produce color fields that
are washed out. Zhang et al.~\cite{ZhangColorful} address this by performing a re-balancing in the loss that 
takes into account the different populations of colors in the training data.
The goal of re-balancing is to give higher weight to rarer colors with respect
to the common colors. 

We adopt a similar strategy that operates in the continuous color field space instead of 
the discrete color field space of Zhang et al. \cite{ZhangColorful}. We use the  
empirical probability estimates (or normalized histogram) $\mathbf{H}$ of colors in the 
quantized `ab' color field computed by \cite{ZhangColorful}. For pixel $p$, we quantize it 
to obtain its bin and retrieve the inverse of probability $\frac{1}{\mathbf{H}_p}$. $\frac{1}{\mathbf{H}_p}$ 
is used as a weight in the squared difference between predicted color $f_{p}(\mathbf{z}, \theta)$ and 
ground-truth $\mathbf{C}_p$ at pixel $p$. Write this loss $\mathcal{L}_{hist}$ in vector form, 

\begin{equation}
\label{eq:vae_loss_hist}
\mathcal{L}_{hist} = \| ({\mathbf{H}^{-1}})^{T} \lbrack \mathbf{C} - f(\mathbf{z}, \theta) \rbrack \|^{2}_{2}
\end{equation}

\noindent
{\bf Gradient.} In addition to the above, we also use a first order loss term that encourages generated 
color fields to have the same gradients as ground truth. Write $\nabla_{h}$ and $\nabla_{v}$
for horizontal and vertical gradient operators. The loss term is,

\begin{equation}
\label{eq:vae_loss_grad}
\mathcal{L}_{grad} = \| \nabla_{h} \mathbf{C} - \nabla_{h} f(\mathbf{z}, \theta) \|^{2}_{2} + \| \nabla_{v} \mathbf{C} - \nabla_{v} f(\mathbf{z}, \theta) \|^{2}_{2}
\end{equation}

Write overall loss $\mathcal{L}_{dec}$ on the decoder as

\begin{equation}
\label{eq:vae_loss_all}
\mathcal{L}_{dec} = \mathcal{L}_{hist} + \lambda_{mah} \mathcal{L}_{mah} + \lambda_{grad} \mathcal{L}_{grad}
\end{equation}

We set hyper-parameters $\lambda_{mah} = .1$ and $\lambda_{grad} = 10^{-3}$. 
The loss on the encoder is the KL-divergence to $\mathcal{N}(0|I)$, same as \cite{AEB}. We
weight this loss by a factor $10^{-2}$ with respect to the decoder loss. This 
relaxes the regularization of the low-dimensional embedding, but gives
greater importance to the fidelity of color field produced by the decoder. 
Our relaxed constraint on embedding space does not have adverse effects.
Because, our conditional model (Refer Section \ref{sec:mdn}) manages to produce 
low-dimensional embeddings which decode to natural colorizations 
(See Figure \ref{fig:divcolor_addl}, \ref{fig:res_cvae_mdn_div}).

\section{Conditional Model (G to z)}
\label{sec:mdn} 

We want to learn a multi-modal (one-to-many) conditional model $P(\mathbf{z}|\mathbf{G})$, 
between the grey-level image $\mathbf{G}$ and the low dimensional embedding $\mathbf{z}$.
Mixture density networks (MDN) model the conditional probability distribution of 
target vectors, conditioned on the input as a mixture of gaussians \cite{BishopMDN}.
This takes into account the one-to-many mapping and allows the target vectors to take 
multiple values conditioned on the same input vector, providing diversity. \\ 

\noindent
{\bf MDN Loss.}
Now, we formulate the loss function for a MDN that models the conditional 
distribution $P(\mathbf{z}|\mathbf{G})$. Here, $P(\mathbf{z}|\mathbf{G})$ is 
Gaussian mixture model with $M$ components. The loss function minimizes the 
conditional negative log likelihood $-\log P(\mathbf{z}|\mathbf{G})$ for
this distribution. Write $\mathcal{L}_{mdn}$ for the MDN loss, $\pi_i$ for the 
mixture coefficients, $\mu_{i}$ for the means and $\sigma$ for the fixed 
spherical co-variance of the GMM. $\pi_i$ and $\mu_i$ are produced 
by a neural network parameterized by $\phi$ with input $\mathbf{G}$. The
MDN loss is, 

\begin{equation}
\label{eq:loss_mdn}
\medmuskip=1mu
\thinmuskip=1mu
\thickmuskip=1mu
\mathcal{L}_{mdn} = - \log P(\mathbf{z} | \mathbf{G}) = - \log \sum\limits_{i=1}^{M} 
\pi_{i}(\mathbf{G}, \phi) \mathcal{N}(\mathbf{z} | \mu_i(\mathbf{G}, \phi), \sigma)  
\end{equation}

It is difficult to optimize \autoref{eq:loss_mdn} since it involves a log of 
summation over exponents of the form 
$e^{\frac{-\|\mathbf{z}-\mu_i(\mathbf{G}, \phi)\|^{2}_{2}}{2\sigma^{2}}}$. The
distance $\| \mathbf{z} - \mu_i(\mathbf{G}, \phi) \|_{2}$ is high when the 
training commences and it leads to a numerical underflow in the exponent.
To avoid this, we pick the gaussian component $m = \arg\min\limits_{i} 
\| \mathbf{z} - \mu_i(\mathbf{G}, \phi) \|_{2}$ with predicted mean closest
to the ground truth code $\mathbf{z}$ and only optimize that component per 
training step. This reduces the loss function to

\begin{equation}
\label{eq:loss_mdn}
\mathcal{L}_{mdn} = - \log \pi_{m}(\mathbf{G}, \phi) + \frac{\|\mathbf{z}-\mu_m(\mathbf{G}, \phi)\|^{2}_{2}}{2\sigma^{2}}
\end{equation}

Intuitively, this $\min$-approximation resolves the identifiability (or symmetry) issue within 
MDN as we tie a grey-level feature to a component ($m^{th}$ component as above). The other components are
free to be optimized by nearby grey-level features. Therefore, clustered
grey-level features jointly optimize the entire GMM, resulting in diverse 
colorizations. In Section \ref{sec:res_div}, we show that this MDN-based strategy 
produces better diverse colorizations than the baseline of CVAE and cGAN
(Section \ref{sec:baseline}). 

\section{Baseline}
\label{sec:baseline}

\noindent
{\bf Conditional Variational Autoencoder (CVAE).} CVAE 
conditions the generative process of VAE on a specific input. 
Therefore, sampling from a CVAE produces diverse outputs for a single 
input. Walker et al. \cite{Walker} use a fully convolutional CVAE for 
diverse motion prediction from a static image. Xue et 
al. \cite{CrossConv} introduce cross-convolutional layers between 
image and motion encoder in CVAE to obtain diverse future frame
synthesis. Zhou and Berg \cite{TLBerg} generate diverse timelapse
videos by incorporating conditional, two-stack and 
recurrent architecture modifications to standard generative models.

Recall that, for our problem of image colorization the input to the CVAE is the 
grey-level image $\mathbf{G}$ and output is the color field $\mathbf{C}$. 
Sohn et al. \cite{Sohn} derive a lower bound on conditional 
log-likelihood $P(\mathbf{C}|\mathbf{G})$ of CVAE. They show that 
CVAE consists of training an encoder $Q(\mathbf{z}|\mathbf{C},\mathbf{G}, \theta)$ 
network with KL-divergence loss and a decoder network $f(\mathbf{z}, \mathbf{G}, \theta)$ with 
an L$_2$ loss. The difference with respect to VAE being that generating the 
embedding and the decoder network both have an additional input $\mathbf{G}$. \\ 

\begin{center}
\begin{figure*}[!t]
\captionsetup[subfigure]{labelformat=empty}
\subfloat{\includegraphics[width=.08\textwidth]{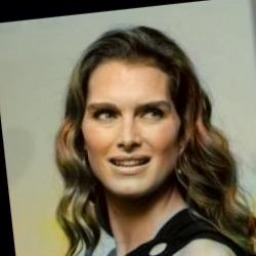}}  
\subfloat{\includegraphics[width=.08\textwidth]{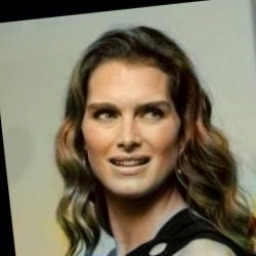}}  
\subfloat[cGAN]{\includegraphics[width=.08\textwidth]{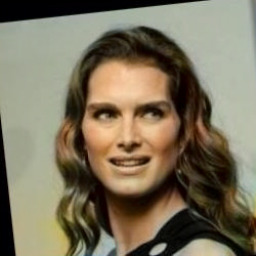}}  
\subfloat{\includegraphics[width=.08\textwidth]{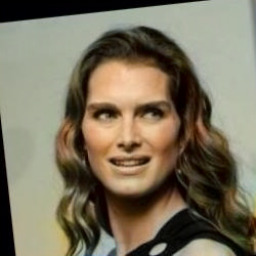}}  
\subfloat{\includegraphics[width=.08\textwidth]{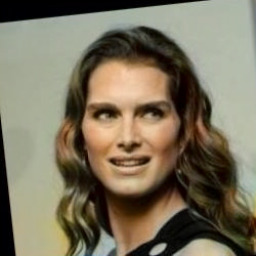}}  \hspace{8pt} 
\subfloat{\includegraphics[width=.08\textwidth]{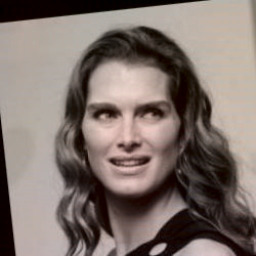}}  
\subfloat{\includegraphics[width=.08\textwidth]{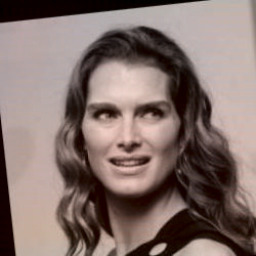}}  
\subfloat[CVAE]{\includegraphics[width=.08\textwidth]{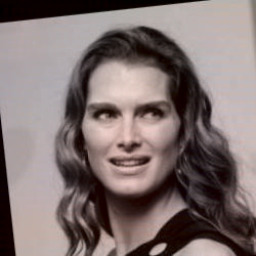}}  
\subfloat{\includegraphics[width=.08\textwidth]{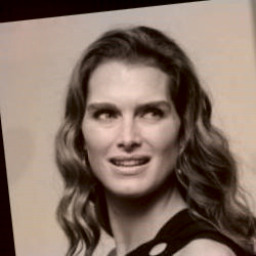}}  
\subfloat{\includegraphics[width=.08\textwidth]{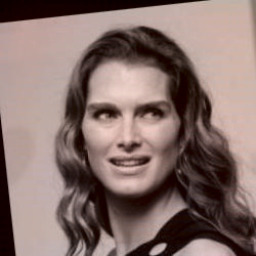}} \hspace{8pt}
\subfloat[GT]{\includegraphics[width=.08\textwidth]{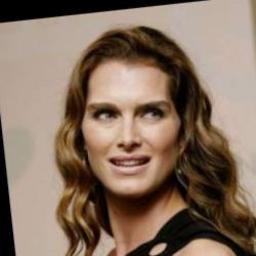}} \\ [-2.5ex]
\subfloat{\includegraphics[width=.08\textwidth]{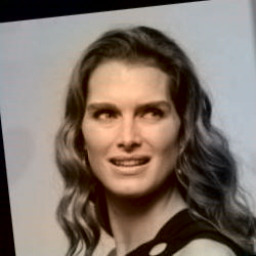}}  
\subfloat{\includegraphics[width=.08\textwidth]{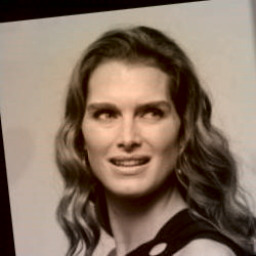}}  
\subfloat[Ours]{\includegraphics[width=.08\textwidth]{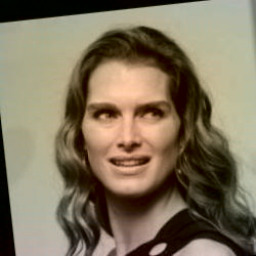}}  
\subfloat{\includegraphics[width=.08\textwidth]{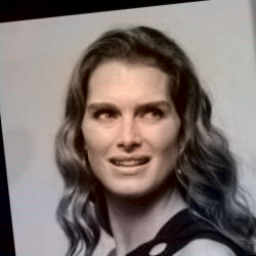}}  
\subfloat{\includegraphics[width=.08\textwidth]{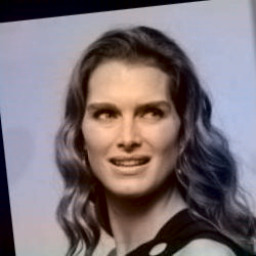}}  \hspace{8pt} 
\subfloat{\includegraphics[width=.08\textwidth]{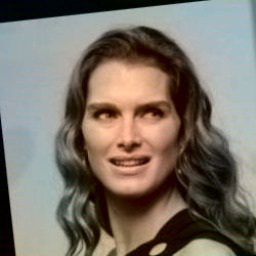}}  
\subfloat{\includegraphics[width=.08\textwidth]{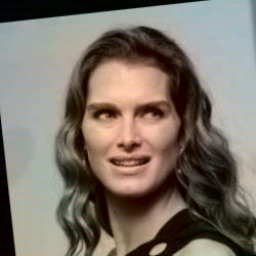}}  
\subfloat[Ours+Skip]{\includegraphics[width=.08\textwidth]{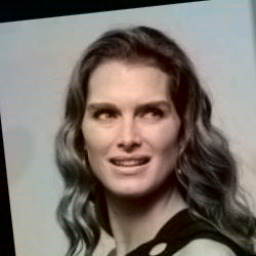}}  
\subfloat{\includegraphics[width=.08\textwidth]{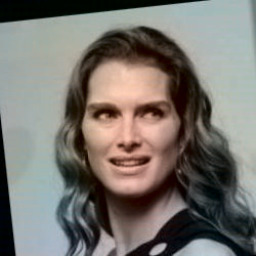}}  
\subfloat{\includegraphics[width=.08\textwidth]{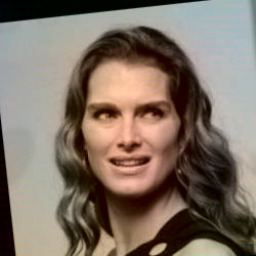}} \\[-2ex]
\subfloat{\includegraphics[width=.08\textwidth]{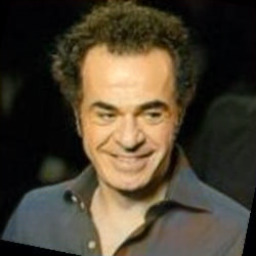}}  
\subfloat{\includegraphics[width=.08\textwidth]{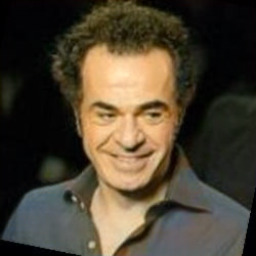}}  
\subfloat[cGAN]{\includegraphics[width=.08\textwidth]{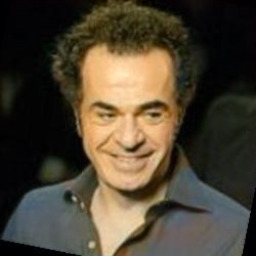}}  
\subfloat{\includegraphics[width=.08\textwidth]{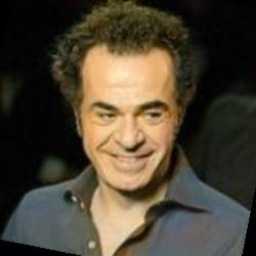}}  
\subfloat{\includegraphics[width=.08\textwidth]{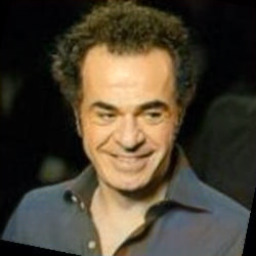}}  \hspace{8pt} 
\subfloat{\includegraphics[width=.08\textwidth]{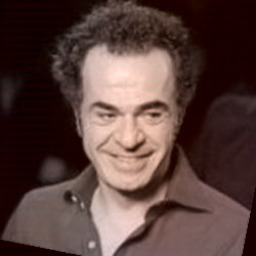}}  
\subfloat{\includegraphics[width=.08\textwidth]{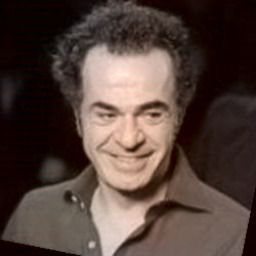}}  
\subfloat[CVAE]{\includegraphics[width=.08\textwidth]{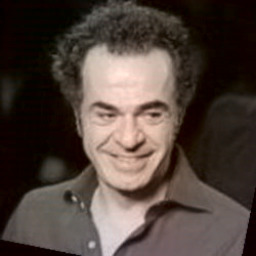}}  
\subfloat{\includegraphics[width=.08\textwidth]{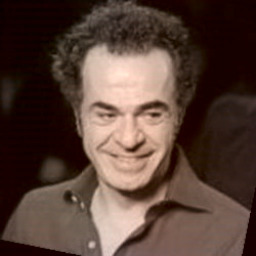}}  
\subfloat{\includegraphics[width=.08\textwidth]{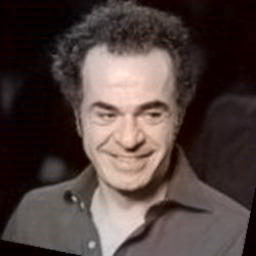}} \hspace{8pt}
\subfloat[GT]{\includegraphics[width=.08\textwidth]{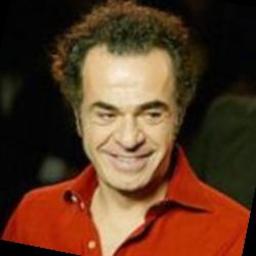}} \\[-2.5ex]
\subfloat{\includegraphics[width=.08\textwidth]{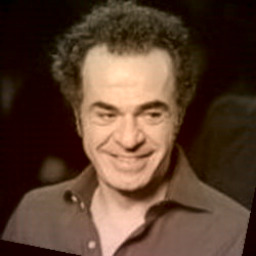}}  
\subfloat{\includegraphics[width=.08\textwidth]{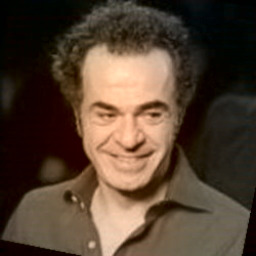}}  
\subfloat[Ours]{\includegraphics[width=.08\textwidth]{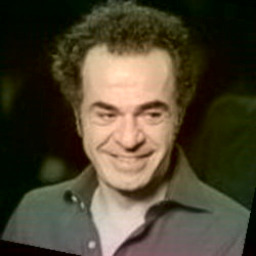}}  
\subfloat{\includegraphics[width=.08\textwidth]{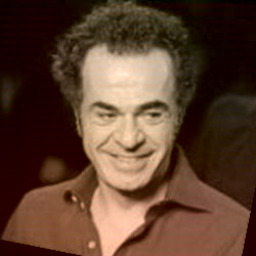}}  
\subfloat{\includegraphics[width=.08\textwidth]{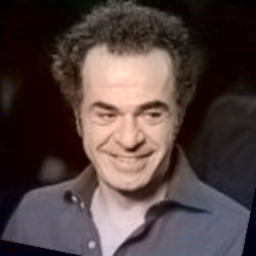}}  \hspace{8pt} 
\subfloat{\includegraphics[width=.08\textwidth]{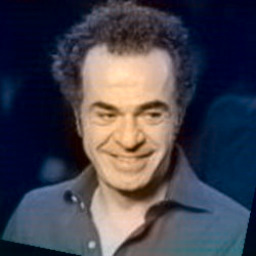}}  
\subfloat{\includegraphics[width=.08\textwidth]{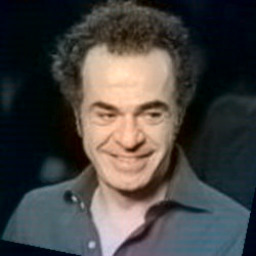}}  
\subfloat[Ours+Skip]{\includegraphics[width=.08\textwidth]{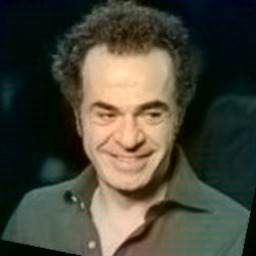}}  
\subfloat{\includegraphics[width=.08\textwidth]{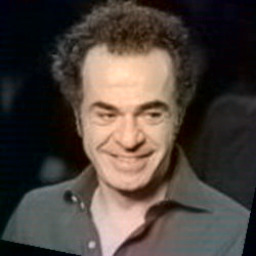}}  
\subfloat{\includegraphics[width=.08\textwidth]{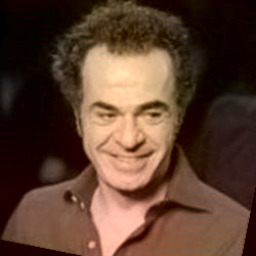}} \\ [-2ex]
\subfloat{\includegraphics[width=.08\textwidth]{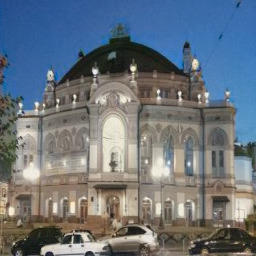}}  
\subfloat{\includegraphics[width=.08\textwidth]{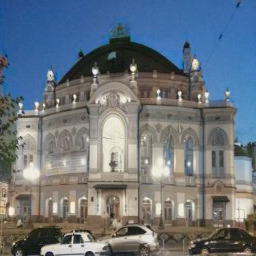}}  
\subfloat[cGAN]{\includegraphics[width=.08\textwidth]{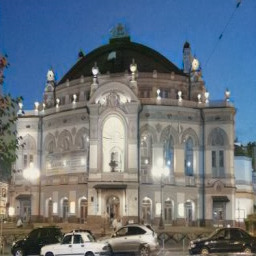}}  
\subfloat{\includegraphics[width=.08\textwidth]{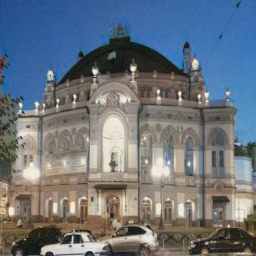}}  
\subfloat{\includegraphics[width=.08\textwidth]{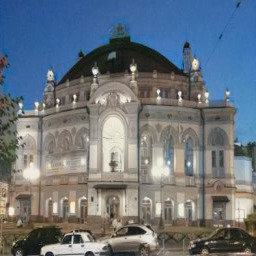}}  \hspace{8pt} 
\subfloat{\includegraphics[width=.08\textwidth]{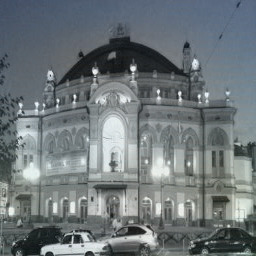}}  
\subfloat{\includegraphics[width=.08\textwidth]{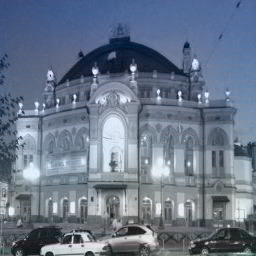}}  
\subfloat[CVAE]{\includegraphics[width=.08\textwidth]{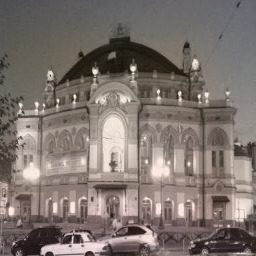}}  
\subfloat{\includegraphics[width=.08\textwidth]{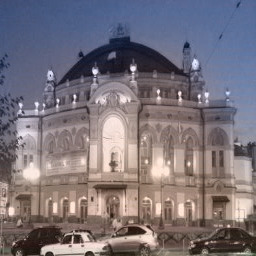}}  
\subfloat{\includegraphics[width=.08\textwidth]{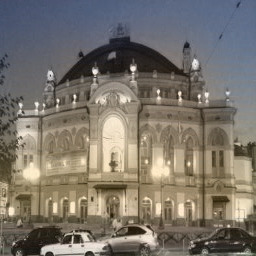}} \hspace{8pt}
\subfloat[GT]{\includegraphics[width=.08\textwidth]{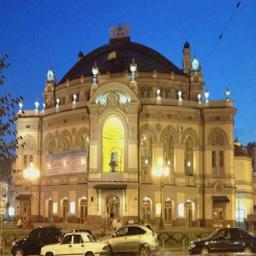}} \\[-2.5ex]
\subfloat{\includegraphics[width=.08\textwidth]{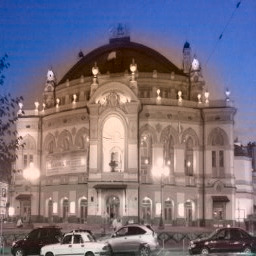}}  
\subfloat{\includegraphics[width=.08\textwidth]{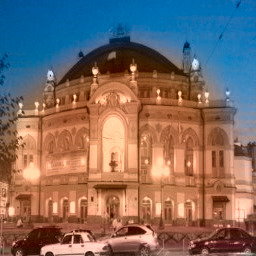}}  
\subfloat[Ours]{\includegraphics[width=.08\textwidth]{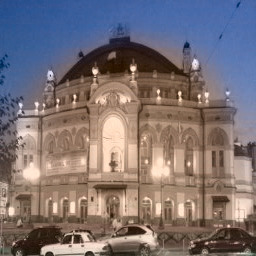}}  
\subfloat{\includegraphics[width=.08\textwidth]{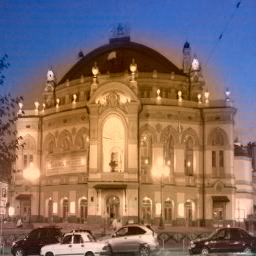}}  
\subfloat{\includegraphics[width=.08\textwidth]{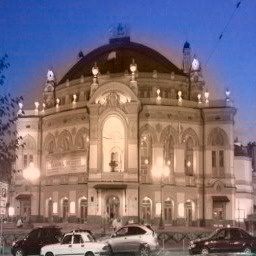}}  \hspace{8pt} 
\subfloat{\includegraphics[width=.08\textwidth]{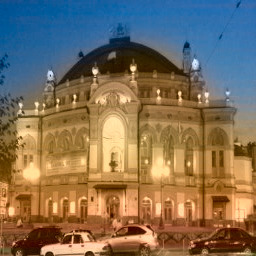}}  
\subfloat{\includegraphics[width=.08\textwidth]{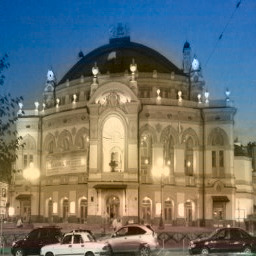}}  
\subfloat[Ours+Skip]{\includegraphics[width=.08\textwidth]{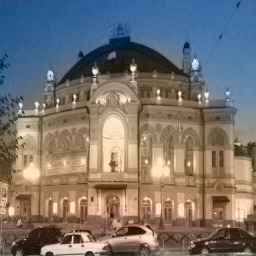}}  
\subfloat{\includegraphics[width=.08\textwidth]{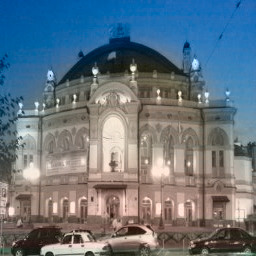}}  
\subfloat{\includegraphics[width=.08\textwidth]{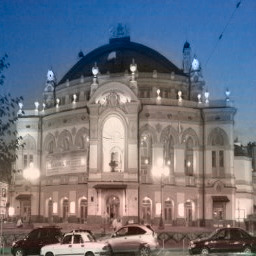}} \\[-2ex]
\subfloat{\includegraphics[width=.08\textwidth]{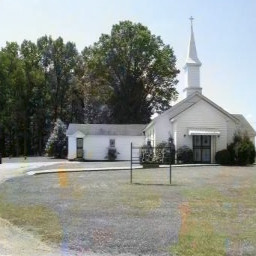}}  
\subfloat{\includegraphics[width=.08\textwidth]{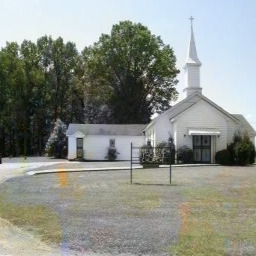}}  
\subfloat[cGAN]{\includegraphics[width=.08\textwidth]{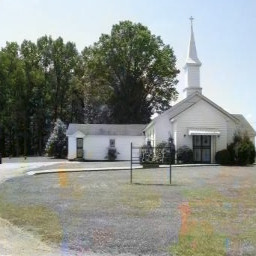}}  
\subfloat{\includegraphics[width=.08\textwidth]{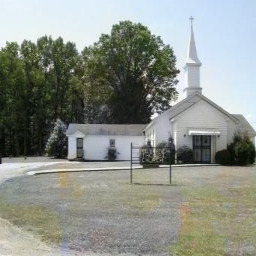}}  
\subfloat{\includegraphics[width=.08\textwidth]{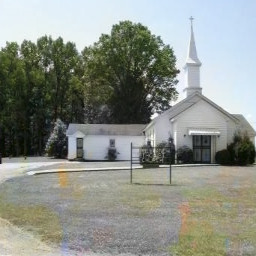}}  \hspace{8pt} 
\subfloat{\includegraphics[width=.08\textwidth]{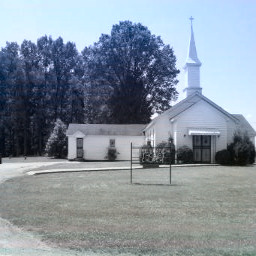}}  
\subfloat{\includegraphics[width=.08\textwidth]{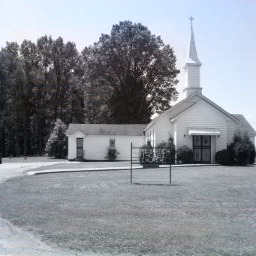}}  
\subfloat[CVAE]{\includegraphics[width=.08\textwidth]{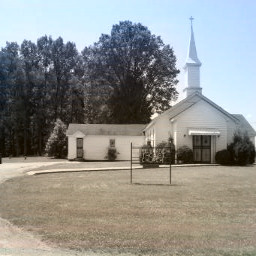}}  
\subfloat{\includegraphics[width=.08\textwidth]{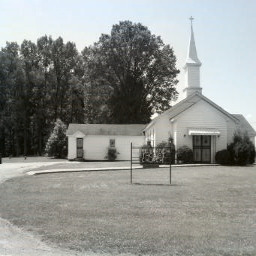}}  
\subfloat{\includegraphics[width=.08\textwidth]{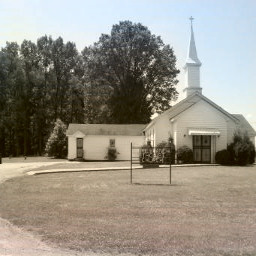}} \hspace{8pt}
\subfloat[GT]{\includegraphics[width=.08\textwidth]{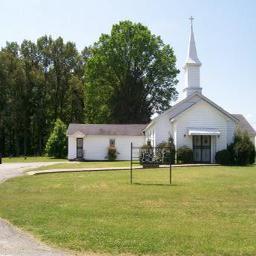}} \\[-2.5ex]
\subfloat{\includegraphics[width=.08\textwidth]{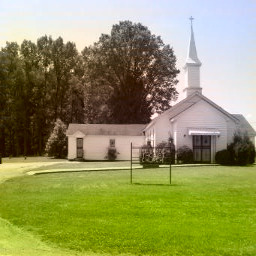}}  
\subfloat{\includegraphics[width=.08\textwidth]{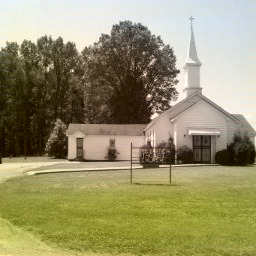}}  
\subfloat[Ours]{\includegraphics[width=.08\textwidth]{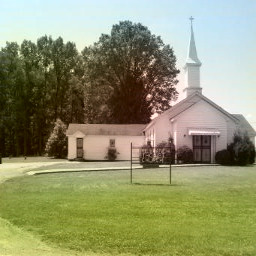}}  
\subfloat{\includegraphics[width=.08\textwidth]{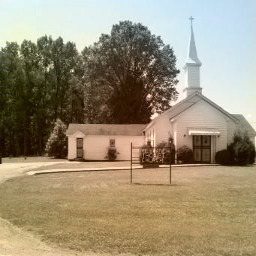}}  
\subfloat{\includegraphics[width=.08\textwidth]{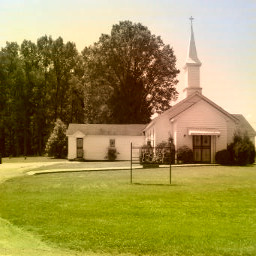}}  \hspace{8pt} 
\subfloat{\includegraphics[width=.08\textwidth]{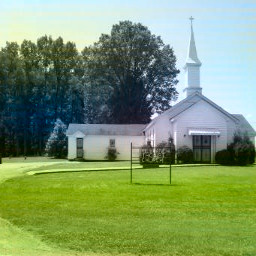}}  
\subfloat{\includegraphics[width=.08\textwidth]{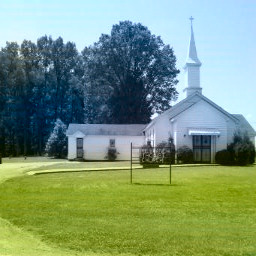}}  
\subfloat[Ours+Skip]{\includegraphics[width=.08\textwidth]{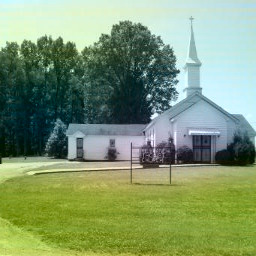}}  
\subfloat{\includegraphics[width=.08\textwidth]{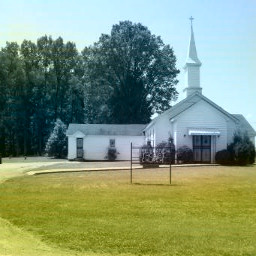}}  
\subfloat{\includegraphics[width=.08\textwidth]{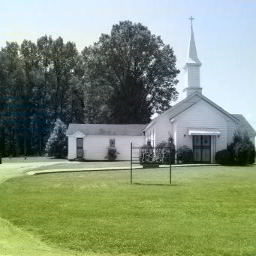}} \\ [-2ex]
\subfloat{\includegraphics[width=.08\textwidth]{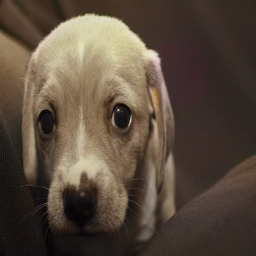}}  
\subfloat{\includegraphics[width=.08\textwidth]{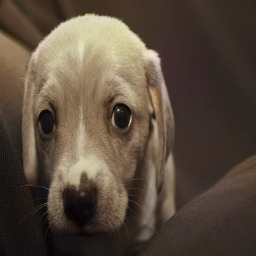}}  
\subfloat[cGAN]{\includegraphics[width=.08\textwidth]{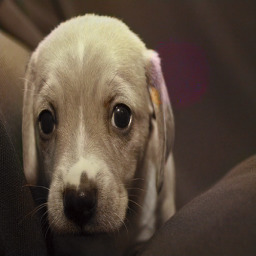}}  
\subfloat{\includegraphics[width=.08\textwidth]{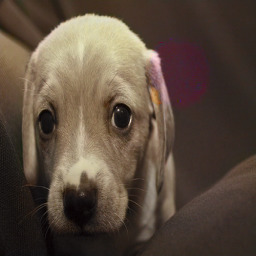}}  
\subfloat{\includegraphics[width=.08\textwidth]{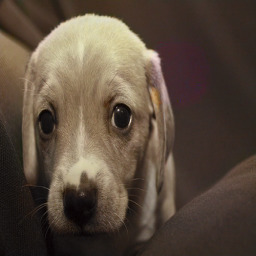}}  \hspace{8pt} 
\subfloat{\includegraphics[width=.08\textwidth]{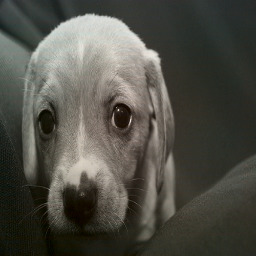}}  
\subfloat{\includegraphics[width=.08\textwidth]{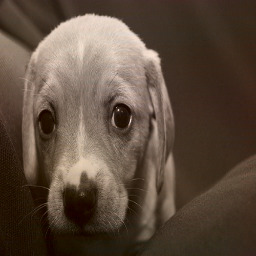}}  
\subfloat[CVAE]{\includegraphics[width=.08\textwidth]{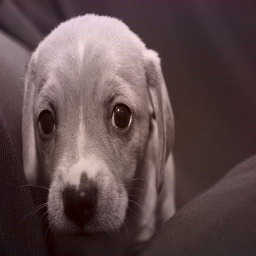}}  
\subfloat{\includegraphics[width=.08\textwidth]{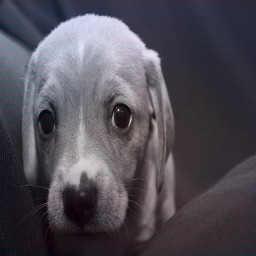}}  
\subfloat{\includegraphics[width=.08\textwidth]{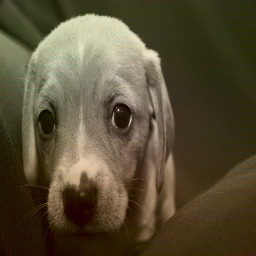}} \hspace{8pt}
\subfloat[GT]{\includegraphics[width=.08\textwidth]{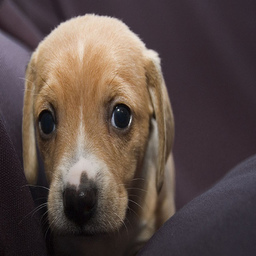}} \\[-2.5ex]
\subfloat{\includegraphics[width=.08\textwidth]{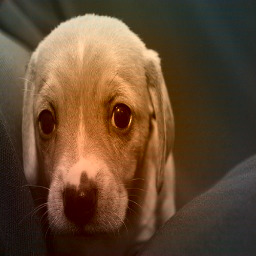}}  
\subfloat{\includegraphics[width=.08\textwidth]{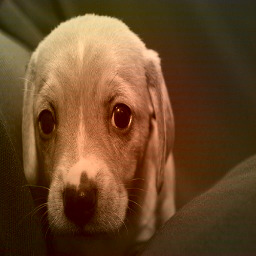}}  
\subfloat[Ours]{\includegraphics[width=.08\textwidth]{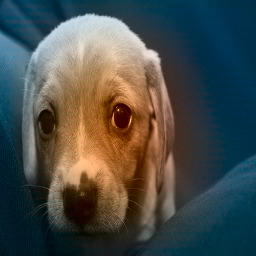}}  
\subfloat{\includegraphics[width=.08\textwidth]{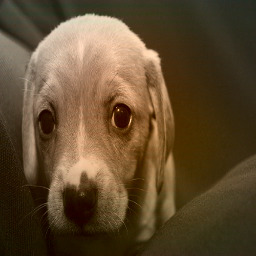}}  
\subfloat{\includegraphics[width=.08\textwidth]{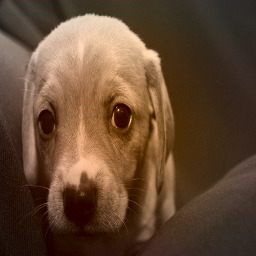}}  \hspace{8pt} 
\subfloat{\includegraphics[width=.08\textwidth]{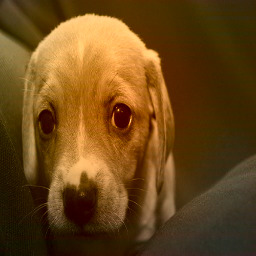}}  
\subfloat{\includegraphics[width=.08\textwidth]{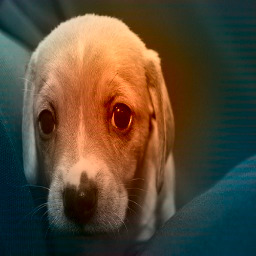}}  
\subfloat[Ours+Skip]{\includegraphics[width=.08\textwidth]{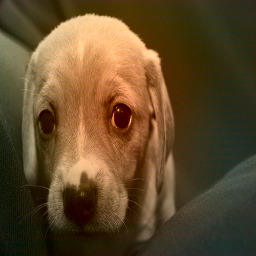}}  
\subfloat{\includegraphics[width=.08\textwidth]{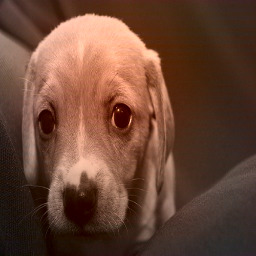}}  
\subfloat{\includegraphics[width=.08\textwidth]{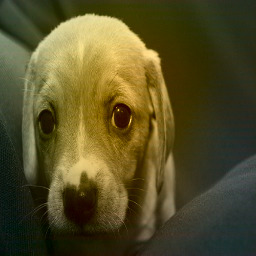}} \\ [-2ex]
\subfloat{\includegraphics[width=.08\textwidth]{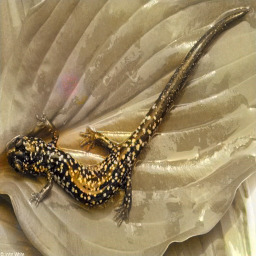}}  
\subfloat{\includegraphics[width=.08\textwidth]{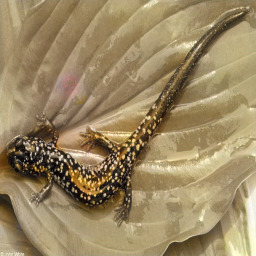}}  
\subfloat[cGAN]{\includegraphics[width=.08\textwidth]{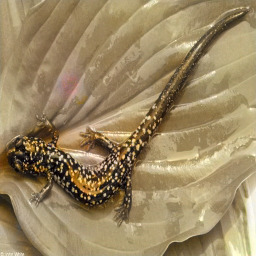}}  
\subfloat{\includegraphics[width=.08\textwidth]{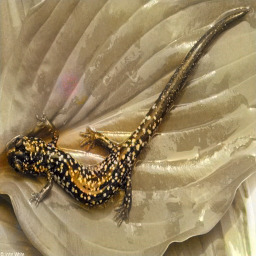}}  
\subfloat{\includegraphics[width=.08\textwidth]{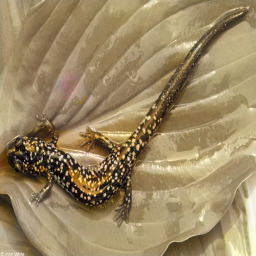}}  \hspace{8pt} 
\subfloat{\includegraphics[width=.08\textwidth]{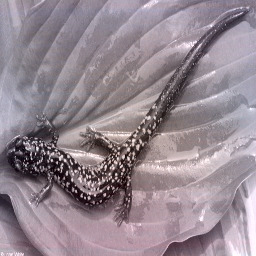}}  
\subfloat{\includegraphics[width=.08\textwidth]{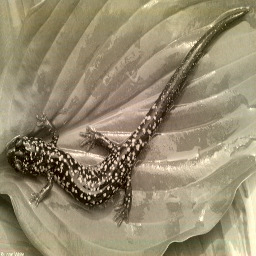}}  
\subfloat[CVAE]{\includegraphics[width=.08\textwidth]{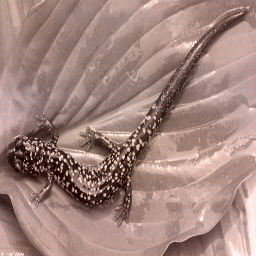}}  
\subfloat{\includegraphics[width=.08\textwidth]{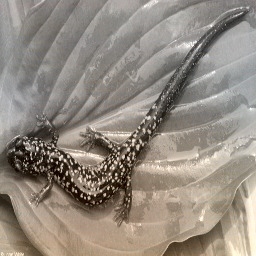}}  
\subfloat{\includegraphics[width=.08\textwidth]{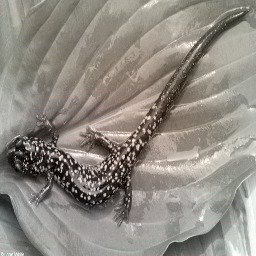}} \hspace{8pt}
\subfloat[GT]{\includegraphics[width=.08\textwidth]{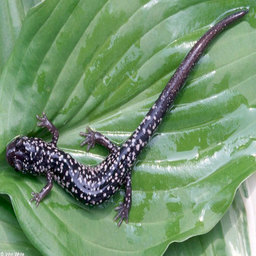}} \\[-2.5ex]
\subfloat{\includegraphics[width=.08\textwidth]{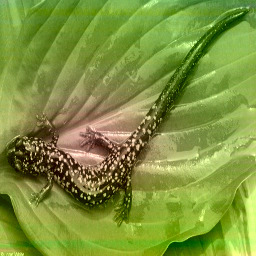}}  
\subfloat{\includegraphics[width=.08\textwidth]{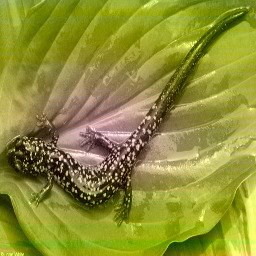}}  
\subfloat[Ours]{\includegraphics[width=.08\textwidth]{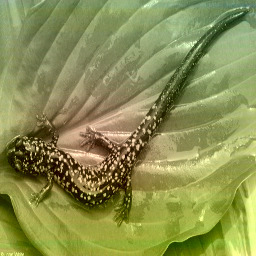}}  
\subfloat{\includegraphics[width=.08\textwidth]{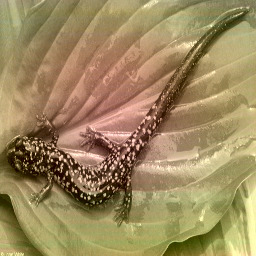}}  
\subfloat{\includegraphics[width=.08\textwidth]{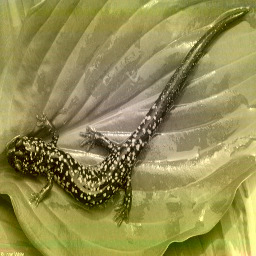}}  \hspace{8pt} 
\subfloat{\includegraphics[width=.08\textwidth]{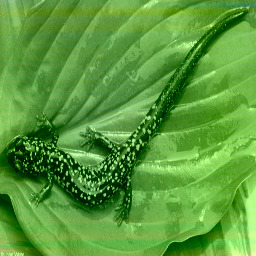}}  
\subfloat{\includegraphics[width=.08\textwidth]{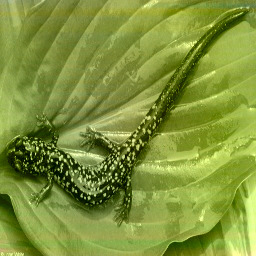}}  
\subfloat[Ours+Skip]{\includegraphics[width=.08\textwidth]{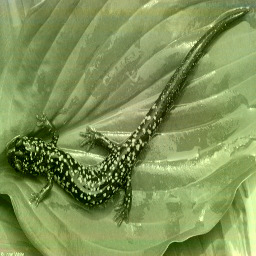}}  
\subfloat{\includegraphics[width=.08\textwidth]{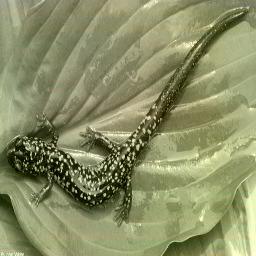}}  
\subfloat{\includegraphics[width=.08\textwidth]{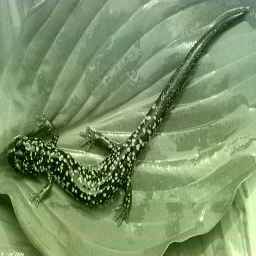}} \\ [-2ex]
\caption{Diverse colorizations from our methods are compared to the CVAE, cGAN and the ground-truth (GT). We can 
generate diverse colorizations, which cGAN \cite{Isola} do not. CVAE colorizations have low diversity and artifacts.} 
\label{fig:res_cvae_mdn_div} 
\end{figure*}
\end{center}

\noindent
{\bf Conditional Generative Adversarial Network (cGAN).} Isola et al.~\cite{Isola} recently
proposed a cGAN based architecture to solve various image-to-image translation tasks.
One of which is colorizing grey-level images. They use an encoder-decoder architecture
along with skip connections that propagate low-level detail. The network is trained
with a patch-based adversarial loss, in addition to L$_{1}$ loss. The noise 
(or embedding $\mathbf{z}$) is provided in the form of dropout~\cite{Srivastava}. 
At test-time, we use dropout to generate diverse colorizations. We cluster
$256$ colorizations into $5$ cluster centers (See cGAN in 
Figure \ref{fig:res_cvae_mdn_div}). \\

\begin{figure}[t]
\centering
\includegraphics[trim={0 9cm 5cm 0},clip,width=.5\textwidth]{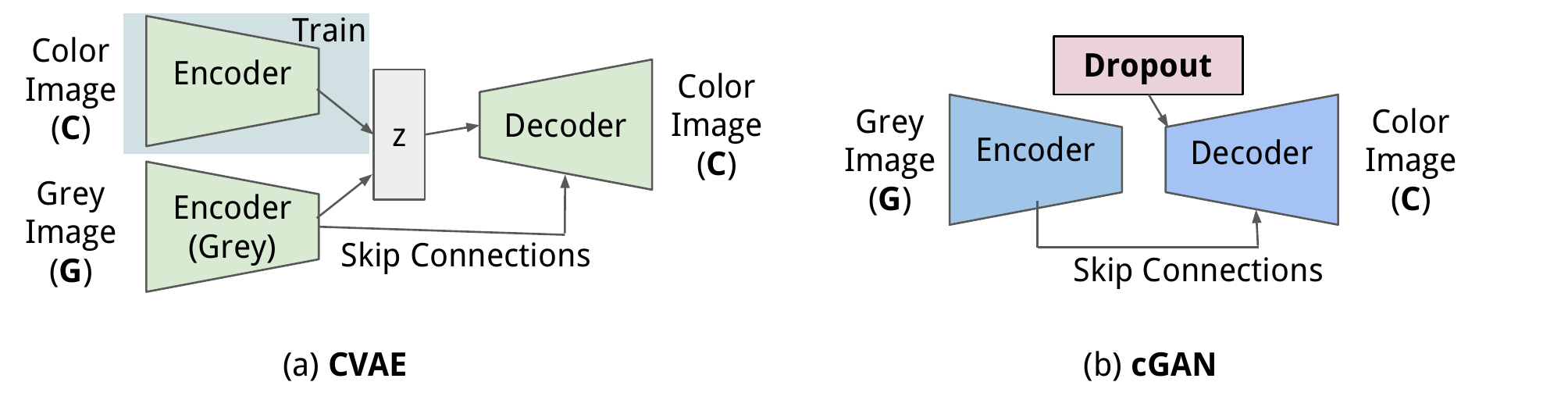}
\caption{Illustration of the CVAE baseline (left) and cGAN baseline (right). For CVAE, the
embedding $\mathbf{z}$ is generated by using both $\mathbf{C}$ and $\mathbf{G}$. The decoder
network is conditioned on $\mathbf{G}$, in addition to $\mathbf{z}$. At test time, we
do not use the highlighted encoder and embedding $\mathbf{z}$ is sampled randomly. 
cGAN consists of an encoder-decoder network with skip connections, and noise or 
embedding is due to dropout.} 
\label{fig:baseline} 
\end{figure}

An illustration of these baseline methods is in Figure~\ref{fig:baseline}. We compare 
CVAE and cGAN to our strategy of using VAE and MDN (Figure \ref{fig:overview}) for the 
problem of diverse colorization (Figure \ref{fig:res_cvae_mdn_div}).

\section{Architecture and Implementation Details}

\noindent
{\bf Notation.} Before we begin describing the network architecture, note
the following notation. Write $C_{a}(k, s, n)$ for convolutions with kernel size 
$k$, stride $s$, output channels $n$ and activation $a$, $B$ for batch 
normalization, $U(f)$ for bilinear up-sampling with scale factor $f$ and 
$F(n)$ for fully connected layer with output channels $n$. Note, we perform
convolutions with zero-padding and our fully connected layers use dropout
regularization \cite{Srivastava}. \\

\begin{figure}[t]
\centering
\includegraphics[trim={0 6cm 8cm 0},clip,width=.5\textwidth]{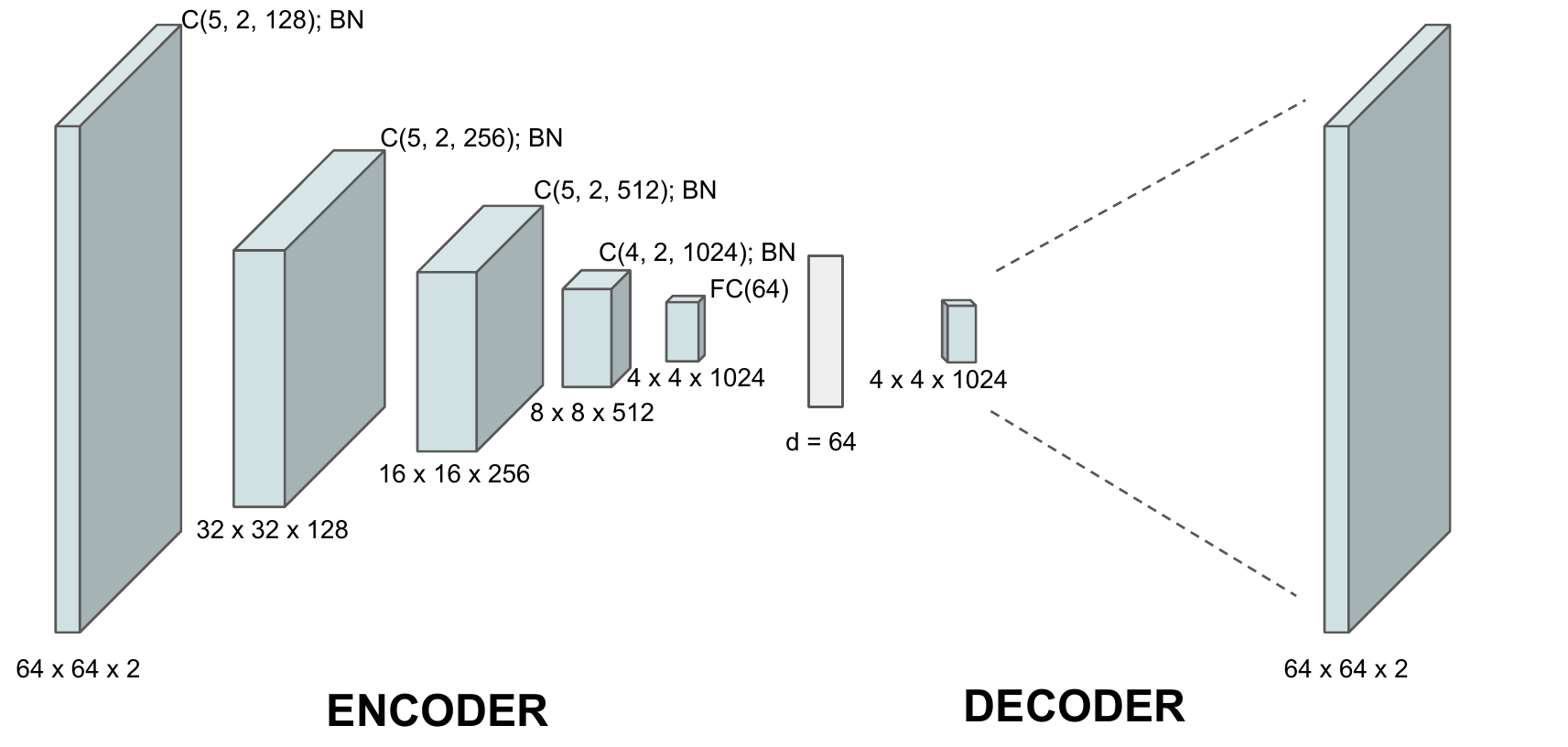}
\caption{An illustration of our VAE architecture. The dimensions of feature maps are at
the bottom and the operations applied to the feature map are indicated at the top. This 
figure shows the encoder. For the decoder architecture, refer to the details in 
Section \ref{sec:arch_vae}.} 
\label{fig:vae_arch} 
\end{figure}

\subsection{VAE}
\label{sec:arch_vae}

Radford et al.\ propose a DCGAN architecture with generator (or decoder) network 
that can model complex spatial structure of images \cite{DCGAN}. We model the 
decoder network of our VAE to be similar to the 
generator network of Radford et al.~\cite{DCGAN}. We follow their best practices of
using strided convolutions instead of pooling, batch normalization \cite{BN}, ReLU activations for 
intermediate layers and tanh for output layer, avoiding fully connected layers
except when decorrelation is required to obtain the low-dimensional embedding. The encoder 
network is roughly the mirror of decoder network, as per the standard practice 
for autoencoder networks. See Figure \ref{fig:vae_arch} for an illustration of our VAE 
architecture.\\

\noindent
{\bf Encoder Network.} The encoder network accepts a color field of size $64 \times 64 \times 2$
and outputs a $d-$dimensional embedding. Encoder network can be written as,  
Input: $64 \times 64 \times 2 \rightarrow C_{ReLU}(5, 2, 128) \rightarrow B \rightarrow C_{ReLU}(5, 2, 256)
\rightarrow B \rightarrow C_{ReLU}(5, 2, 512) \rightarrow B \rightarrow C_{ReLU}(4, 2, 1024) 
\rightarrow B \rightarrow F(d)$. \\

\noindent
{\bf Decoder Network.} The decoder network accepts a $d$-dimensional embedding.
It performs 5 operations of bilinear up-sampling and convolutions to finally output a 
$64 \times 64 \times 2$ color field (a and b of Lab color space comprise the two 
output channels). The decoder network can be written as, 
Input: $1 \times 1 \times d \rightarrow U(4) \rightarrow C_{ReLU}(4, 1, 1024) \rightarrow
B \rightarrow U(2) \rightarrow C_{ReLU}(5, 1, 512) \rightarrow B \rightarrow U(2) 
\rightarrow C_{ReLU}(5, 1, 256) \rightarrow B \rightarrow U(2) \rightarrow C_{ReLU}(5, 1, 128) \rightarrow B 
\rightarrow U(2) \rightarrow C_{tanh}(5, 1, 2)$.

We use $d=64$ for all our three datasets (Section \ref{sec:res_datasets}).

\subsection{MDN}
\label{sec:arch_mdn}
The input to MDN are the grey-level features $\mathbf{G}$ from \cite{ZhangColorful}
and have dimension $28 \times 28 \times 512$. We use $8$ components in the output 
GMM of MDN. The output layer comprises $8 \times d$ activations for means and $8$ 
softmax-ed activations for mixture weights of the $8$ components. We use a fixed spherical variance of 
$.1$. The MDN network
uses 5 convolutional layers followed by two fully connected layers and
can be written as, Input: $28 \times 28 \times 512 \rightarrow C_{ReLU}(5, 1, 384) \rightarrow B 
\rightarrow C_{ReLU}(5, 1, 320) \rightarrow B \rightarrow C_{ReLU}(5, 1, 288) \rightarrow B 
\rightarrow C_{ReLU}(5, 2, 256) \rightarrow B \rightarrow C_{ReLU}(5, 1, 128) \rightarrow B 
\rightarrow FC(4096) \rightarrow FC(8 \times d + 8)$. Equivalently, the MDN
is a network with $12$ convolutional and 2 fully connected layers, with the first $7$ 
convolutional layers pre-trained on task of \cite{ZhangColorful} and held fixed.

At test time, we can sample multiple embeddings from MDN and then generate diverse 
colorizations using VAE decoder. However, to study diverse colorizations in a principled 
manner we adopt a different procedure. We order the predicted means $\mu_i$ in descending 
order of mixture weights $\pi_i$ and use these top-k ($k=5$) means as diverse colorizations
shown in Figure \ref{fig:res_cvae_mdn_div} (See ours, ours+skip).

\subsection{CVAE}
\label{sec:arch_cvae}
In CVAE, the encoder and the decoder both take an additional input $\mathbf{G}$. 
We need an encoder for grey-level images as shown in Figure \ref{fig:baseline}.  The
color image encoder and the decoder are same as the VAE (Section \ref{sec:arch_vae}). The grey-level 
encoder of CVAE can be written as, Input: $64 \times 64 \rightarrow C_{ReLU}(5, 2, 128) \rightarrow B 
\rightarrow C_{ReLU}(5, 2, 256) \rightarrow B \rightarrow C_{ReLU}(5, 2, 512) \rightarrow B 
\rightarrow C_{ReLU}(4, 2, d)$. This produces an output feature map of $4 \times 4 \times d$. 
The $d$-dimensional latent variable generated by the VAE (or color) encoder is spatially 
replicated ($4 \times 4$) and multiplied to the output of grey-level encoder, which forms
the input to the decoder. Additionally, we add skip connections from the grey-level encoder to the 
decoder similar to \cite{Isola}.

At test time, we feed multiple embeddings (randomly sampled) to the CVAE decoder along with 
fixed grey-level input. We feed $256$ embeddings and 
cluster outputs to $5$ colorizations (See CVAE in Figure \ref{fig:res_cvae_mdn_div}). 

Refer to \url{http://vision.cs.illinois.edu/projects/divcolor} for our tensorflow code.

\section{Results}

\begin{table*}[!t]
\begin{tabular}{m{.05\textwidth}m{.095\textwidth}m{.095\textwidth}m{.095\textwidth}m{.095\textwidth}m{.095\textwidth}m{.095\textwidth}m{.095\textwidth}m{.095\textwidth}}
\centering
L$_2$ Loss &
\subfloat{\includegraphics[width=.0950\textwidth]{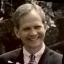}}   & 
\subfloat{\includegraphics[width=.0950\textwidth]{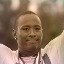}}   &
\subfloat{\includegraphics[width=.0950\textwidth]{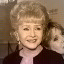}}   &
\subfloat{\includegraphics[width=.0950\textwidth]{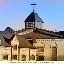}}   & 
\subfloat{\includegraphics[width=.0950\textwidth]{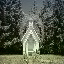}}   &
\subfloat{\includegraphics[width=.0950\textwidth]{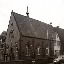}}   &
\subfloat{\includegraphics[width=.0950\textwidth]{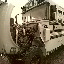}}   & 
\subfloat{\includegraphics[width=.0950\textwidth]{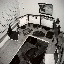}}   \\
Only $\mathcal{L}_{mah}$ &
\subfloat{\includegraphics[width=.0950\textwidth]{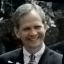}}   & 
\subfloat{\includegraphics[width=.0950\textwidth]{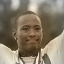}}   &
\subfloat{\includegraphics[width=.0950\textwidth]{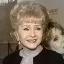}}   &
\subfloat{\includegraphics[width=.0950\textwidth]{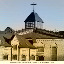}}   & 
\subfloat{\includegraphics[width=.0950\textwidth]{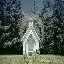}}   &
\subfloat{\includegraphics[width=.0950\textwidth]{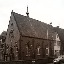}}   &
\subfloat{\includegraphics[width=.0950\textwidth]{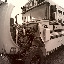}}   &
\subfloat{\includegraphics[width=.0950\textwidth]{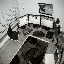}}   \\ 
All Terms&
\subfloat{\includegraphics[width=.0950\textwidth]{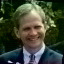}}  & 
\subfloat{\includegraphics[width=.0950\textwidth]{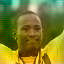}}   &
\subfloat{\includegraphics[width=.0950\textwidth]{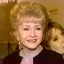}}   &
\subfloat{\includegraphics[width=.0950\textwidth]{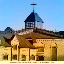}} &  
\subfloat{\includegraphics[width=.0950\textwidth]{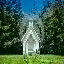}}  & 
\subfloat{\includegraphics[width=.0950\textwidth]{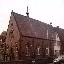}}   &
\subfloat{\includegraphics[width=.0950\textwidth]{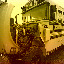}} &   
\subfloat{\includegraphics[width=.0950\textwidth]{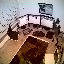}}   \\ 
Ground Truth &
\subfloat{\includegraphics[width=.0950\textwidth]{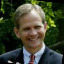}}   &
\subfloat{\includegraphics[width=.0950\textwidth]{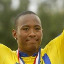}}   &
\subfloat{\includegraphics[width=.0950\textwidth]{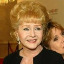}}   &
\subfloat{\includegraphics[width=.0950\textwidth]{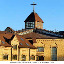}} &  
\subfloat{\includegraphics[width=.0950\textwidth]{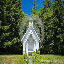}}  & 
\subfloat{\includegraphics[width=.0950\textwidth]{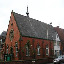}}   &
\subfloat{\includegraphics[width=.0950\textwidth]{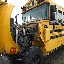}} &   
\subfloat{\includegraphics[width=.0950\textwidth]{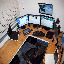}}   \\
\multicolumn{1}{c|}{} & \multicolumn{3}{c|}{LFW} & \multicolumn{3}{c|}{LSUN Church} & \multicolumn{2}{c|}{ImageNet-Val} \\  
\end{tabular}
\captionof{figure}[]{Qualitative results with different loss terms for the VAE decoder network. Top or $1^{st}$ Row uses only the L$_2$ 
loss, $2^{nd}$ row uses $\mathcal{L}_{mah}$, $3^{rd}$ row uses all the loss terms: mahalanobis, colorfulness and gradient 
(See $\mathcal{L}_{dec}$ of \autoref{eq:vae_loss_all}) and last row is the ground-truth color field. These qualitative
results show that using all our loss terms generates better quality color fields as compared to the standard L$_2$ loss for VAE decoders.} 
\label{fig:res_dec_loss} 
\end{table*}

In Section \ref{sec:res_vae_loss}, we evaluate the performance improvement
by the loss terms we construct for the VAE decoder. Section \ref{sec:res_div}
shows the diverse colorizations obtained by our method and we  compare 
it to the CVAE and the cGAN. We also demonstrate the performance of another 
variant of our method: ``ours+skip''. In ours+skip, we use a VAE with 
an additional grey-level encoder and skip connections to the decoder (similar to 
cGAN in Figure \ref{fig:baseline}) and the MDN step is the same. The grey-level 
encoder architecture is the same as CVAE described above.

\begin{table}[!t]
\begin{center}
\setlength{\tabcolsep}{3pt}
\begin{tabular}{|c|c|c|c|c|c|c|}
\specialrule{.1em}{.05em}{.05em}
Dataset & \multicolumn{2}{c|}{L2-Loss} & \multicolumn{2}{c|}{Mah-Loss} & 
\multicolumn{2}{c|}{\specialcell{Mah-Loss \\ + Colorfulness \\ + Gradient}} \\ \cline{2-7}
& All & Grid & All & Grid & All & Grid \\ 
\specialrule{.1em}{.05em}{.05em}
LFW & .034 & .035 & .034 & .032 & {\bf .029} & {\bf .029} \\
Church & .024 & .025 & .026& .026 & {\bf .023} & {\bf .023} \\
\specialcell{ImageNet\\-Val} & {\bf .031} & {\bf .031} & .039 & .039 & .039 & .039 \\
\specialrule{.1em}{.05em}{.05em} 
\end{tabular}
\end{center}
\caption{For test set, our loss terms show better mean absolute error 
per pixel (wrt ground-truth color field) when compared to the standard L$_2$ loss 
on LFW and Church.}
\label{tab:res_l2}
\end{table}

\begin{table}[!t]
\setlength{\tabcolsep}{3pt}
\begin{center}
\begin{tabular}{|c|c|c|c|c|c|c|}
\specialrule{.1em}{.05em}{.05em}
Dataset & \multicolumn{2}{c|}{L2-Loss} & \multicolumn{2}{c|}{Mah-Loss} & 
\multicolumn{2}{c|}{\specialcell{Mah-Loss \\ + Colorfulness \\ + Gradient}} \\ \cline{2-7}
& All & Grid & All & Grid & All & Grid \\ 
\specialrule{.1em}{.05em}{.05em}
LFW & 7.20 & 11.29 & 6.69 & 7.33 &  {\bf 2.65} & {\bf 2.83} \\
Church & 4.9 &  4.68 &  6.54 &  6.42 &  {\bf 1.74} &  {\bf 1.71} \\
\specialcell{ImageNet\\-Val} &  10.02 &  9.21 &  12.99 &  12.19 &  {\bf 4.82} &  {\bf 4.66} \\
\specialrule{.1em}{.05em}{.05em} 
\end{tabular}
\end{center}
\caption{For test set, our loss terms show better weighted absolute error
per pixel (wrt ground-truth color fields) when compared to L$_2$ loss on
all the datasets. Note, having lower weighted error implies, in addition
to common colors, the rarer colors are also predicted correctly. This
implies a higher quality colorization, one that is not washed out.}
\label{tab:res_chi}
\end{table}
 
\subsection{Datasets} 
\label{sec:res_datasets}

We use three datasets with varying complexity of color fields. First, 
we use the Labelled Faces in the Wild dataset (LFW) \cite{Lfw} which consists 
of $13, 233$ face images aligned by deep funneling \cite{deepfun}. 
Since the face images are aligned, this dataset has some structure to it. 
Next, we use the LSUN-Church \cite{lsun} dataset with $126, 227$ images. 
These images are not aligned and lack the structure that was present in the LFW
dataset. They are however images of the same scene category and therefore, 
they are more structured than the images in the wild. Finally,
we use the validation set of ILSVRC-2015 \cite{ILSVRC15} (called ImageNet-Val) 
with $50, 000$ images as our third dataset. These images are the most un-structured of 
the three datasets. For each dataset, we randomly choose a subset of
$1000$ images as test set and use the remaining images for training.

\begin{table}[!t]
\begin{center}
\resizebox{.5\textwidth}{!}{
\begin{tabular}{|c|c|c|c|c|c|c|}
\specialrule{.1em}{.05em}{.05em}
Method & \multicolumn{2}{c|}{LFW} & \multicolumn{2}{c|}{Church} & 
\multicolumn{2}{c|}{ImageNet-Val} \\ \cline{2-7}
& Eob. & Var. & Eob. & Var. & Eob. & Var. \\ 
\specialrule{.1em}{.05em}{.05em}
CVAE &  .031 & $1.0 \times10^{-4}$ &  {\bf .029}  & $2.2 \times10^{-4}$ &   {\bf .037} &  $2.5 \times10^{-4}$\\
cGAN &  .047 & $8.4 \times10^{-6}$ &  .048 &  $6.2 \times10^{-6}$ &   .048 & $8.88 \times 10^{-6}$ \\
Ours &  {\bf .030} & $\mathbf{1.1 \times10^{-3}}$ &  .036 & $\mathbf{3.1 \times10^{-4}}$  &   .043 & $\mathbf{6.8 \times10^{-4}}$ \\
\specialcell{Ours+\\skip} &  .031 &  $4.4 \times10^{-4}$ & .036 & $2.9 \times10^{-4}$&   .041 &  $6.0 \times10^{-4}$\\
\specialrule{.1em}{.05em}{.05em} 
\end{tabular}
}
\end{center}
\caption{For every dataset, we obtain high variance (proxy measure for 
diversity) and often low error-of-best per pixel (Eob.) to the ground-truth using our method. 
This shows our methods generate color fields closer to the ground-truth 
with more diversity compared to the baseline.} 
\label{fig:quant_div} 
\end{table}

\subsection{Effect of Loss terms on VAE Decoder}
\label{sec:res_vae_loss}

We train VAE decoders with: $(i)$ the standard L$_2$ loss, $(ii)$ 
the specificity loss $\mathcal{L}_{mah}$ of Section \ref{sec:vae_dec_loss}, 
and $(iii)$ all our loss terms of \autoref{eq:vae_loss_all}. Figure 
\ref{fig:res_dec_loss} shows the colorizations obtained for the
test set with these different losses. To achieve this
colorization we sample the embedding from the encoder network. 
Therefore, this does not comprise a true colorization task. 
However, it allows us to evaluate the performance of the decoder 
network when the best possible embedding is available. 
Figure \ref{fig:res_dec_loss} shows that the colorizations obtained
with the L$_2$ loss are greyish. In contrast, by using all our loss 
terms we obtain plausible and realistic colorizations with vivid colors. Note the
yellow shirt and the yellow equipment, brown desk and the green trees 
in third row of Figure \ref{fig:res_dec_loss}. For all datasets, using 
all our loss terms provides better colorizations compared to the
standard L$_2$ loss. Note, the face images in 
the second row have more contained skin colors as compared to the first 
row. This shows the subtle benefits obtained from the specificity loss.

In Table \ref{tab:res_l2}, we compare the mean absolute error per-pixel with respect to 
the ground-truth for different loss terms. And, in Table \ref{tab:res_chi},
we compare the mean weighted absolute error per-pixel for these
loss terms. The weighted error uses the same weights as colorfulness loss of 
Section \ref{sec:vae_dec_loss}. We compute the error over: $1)$ all pixels 
(All) and $2)$ over a $8 \times 8$ uniformly spaced grid in the center of 
image (Grid). We compute error on a grid to avoid using too many correlated 
neighboring pixels. On the absolute error metric of Table \ref{tab:res_l2}, 
for LFW and Church, we obtain lower errors with all loss terms as compared to the 
standard L$_2$ loss. Note unlike L$_2$ loss, we do not specifically train for 
this absolute error metric and yet achieve reasonable performance with our 
loss terms. On the weighted error metric of Table \ref{tab:res_chi}, our loss 
terms outperform the standard L$_2$ loss on all datasets.

\subsection{Comparison to baseline}
\label{sec:res_div}

\begin{center}
\begin{figure}[!t]
\captionsetup[subfigure]{labelformat=empty}
\subfloat{\includegraphics[width=.075\textwidth]{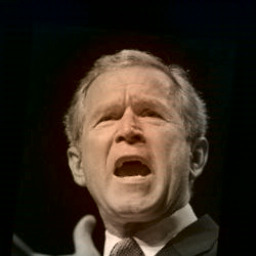}}  
\subfloat{\includegraphics[width=.075\textwidth]{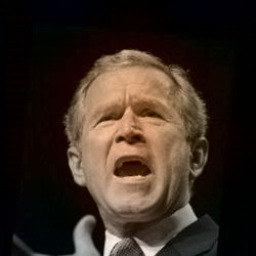}}  
\subfloat{\includegraphics[width=.075\textwidth]{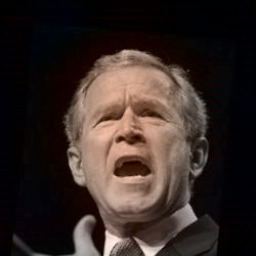}}  
\subfloat{\includegraphics[width=.075\textwidth]{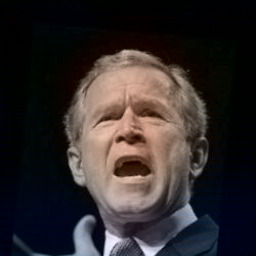}}  
\subfloat{\includegraphics[width=.075\textwidth]{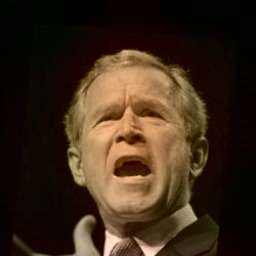}}  \hspace{8pt} 
\subfloat{\includegraphics[width=.075\textwidth]{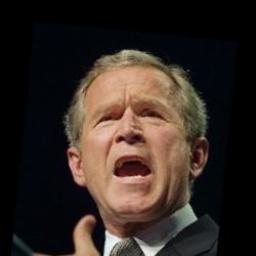}} \\ [-2ex]
\subfloat{\includegraphics[width=.075\textwidth]{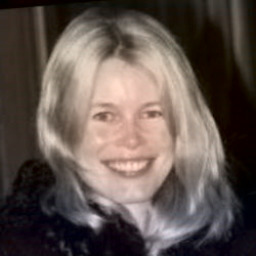}}  
\subfloat{\includegraphics[width=.075\textwidth]{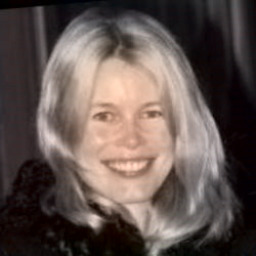}}  
\subfloat{\includegraphics[width=.075\textwidth]{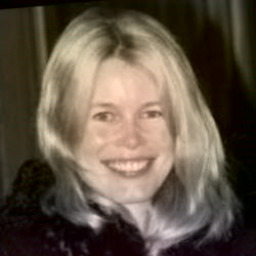}}  
\subfloat{\includegraphics[width=.075\textwidth]{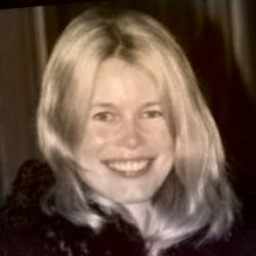}}  
\subfloat{\includegraphics[width=.075\textwidth]{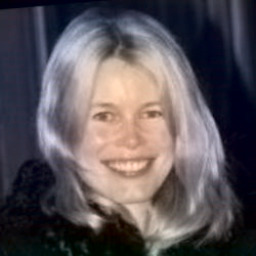}}  \hspace{8pt} 
\subfloat{\includegraphics[width=.075\textwidth]{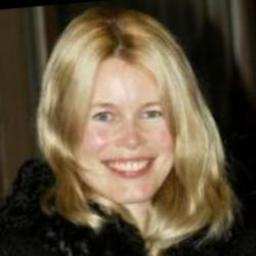}} \\ [-2ex]
\subfloat{\includegraphics[width=.075\textwidth]{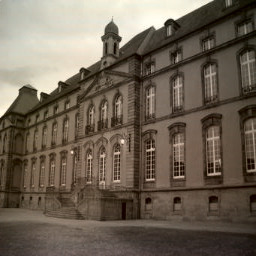}}  
\subfloat{\includegraphics[width=.075\textwidth]{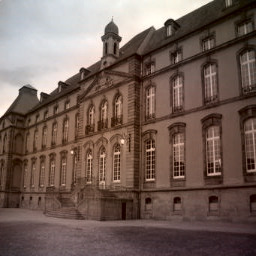}}  
\subfloat{\includegraphics[width=.075\textwidth]{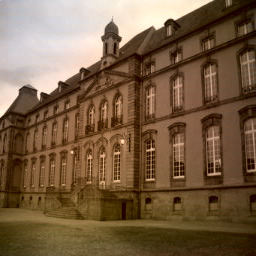}}  
\subfloat{\includegraphics[width=.075\textwidth]{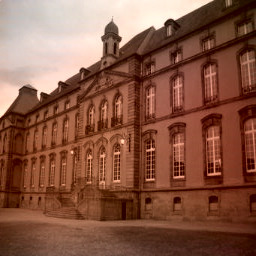}}  
\subfloat{\includegraphics[width=.075\textwidth]{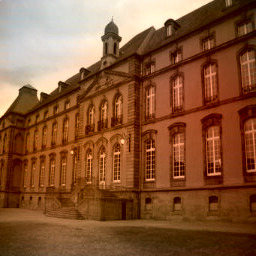}}  \hspace{8pt} 
\subfloat{\includegraphics[width=.075\textwidth]{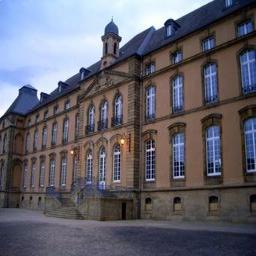}} \\ [-2ex]
\subfloat{\includegraphics[width=.075\textwidth]{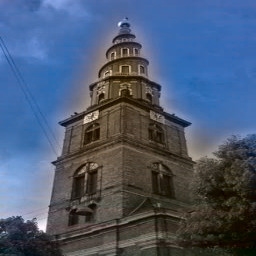}}  
\subfloat{\includegraphics[width=.075\textwidth]{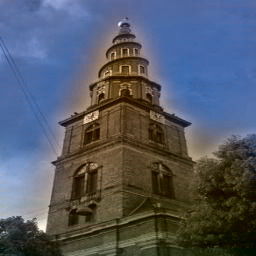}}  
\subfloat{\includegraphics[width=.075\textwidth]{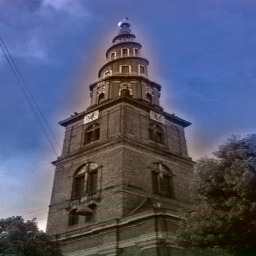}}  
\subfloat{\includegraphics[width=.075\textwidth]{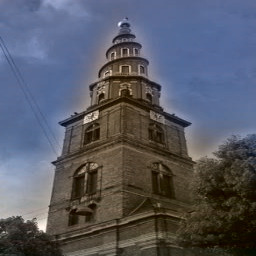}}  
\subfloat{\includegraphics[width=.075\textwidth]{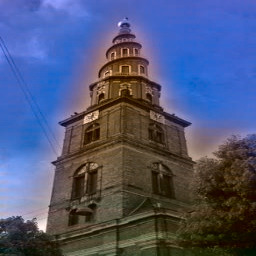}}  \hspace{8pt} 
\subfloat{\includegraphics[width=.075\textwidth]{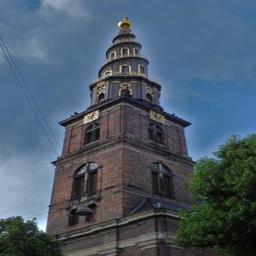}} \\ [-2ex]
\subfloat{\includegraphics[width=.075\textwidth]{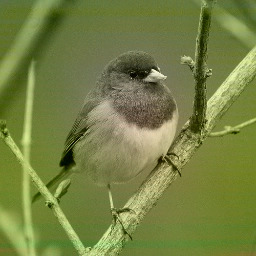}}  
\subfloat{\includegraphics[width=.075\textwidth]{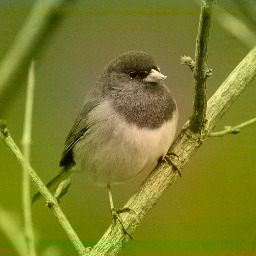}}  
\subfloat{\includegraphics[width=.075\textwidth]{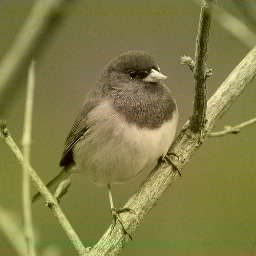}}  
\subfloat{\includegraphics[width=.075\textwidth]{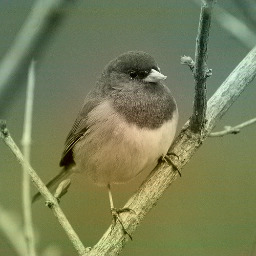}}  
\subfloat{\includegraphics[width=.075\textwidth]{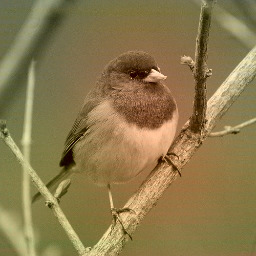}}  \hspace{8pt} 
\subfloat{\includegraphics[width=.075\textwidth]{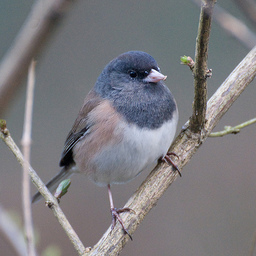}} \\ [-2ex]
\subfloat{\includegraphics[width=.075\textwidth]{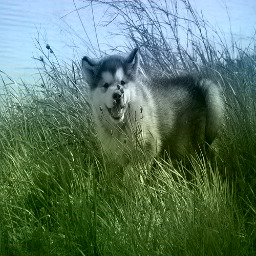}}  
\subfloat{\includegraphics[width=.075\textwidth]{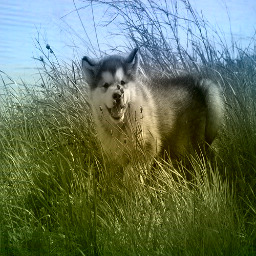}}  
\subfloat[Ours]{\includegraphics[width=.075\textwidth]{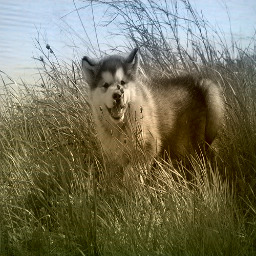}}  
\subfloat{\includegraphics[width=.075\textwidth]{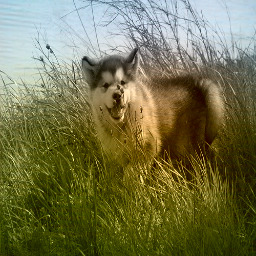}}  
\subfloat{\includegraphics[width=.075\textwidth]{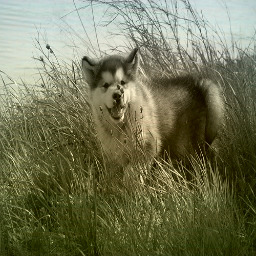}}  \hspace{8pt} 
\subfloat[GT]{\includegraphics[width=.075\textwidth]{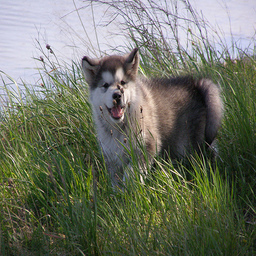}} \\ [-2ex]
\caption{Diverse colorizations from our method. Top two rows are LFW, next two LSUN Church
and last two ImageNet-Val. See Figure \ref{fig:res_cvae_mdn_div} for comparisons to baseline.}
\label{fig:divcolor_addl}
\end{figure}
\end{center}

In Figure \ref{fig:res_cvae_mdn_div}, we compare the diverse colorizations generated
by our strategy (Sections \ref{sec:vae}, \ref{sec:mdn}) and the baseline 
methods -- CVAE and cGAN (Section \ref{sec:baseline}). Qualitatively, we observe that our 
strategy generates better quality diverse colorizations which are each, realistic.
Note that for each dataset, different methods use the same train/test split and 
we train them for $10$ epochs. The diverse colorizations have good quality for
LFW and LSUN Church. We observe different skin tones, hair, cloth and background colors 
for LFW, and we observe different brick, sky and grass colors for LSUN Church.
More colorizations in Figures \ref{fig:divcolor_addl},\ref{fig:res_lfw},\ref{fig:res_church} and \ref{fig:res_imagenetval}.

In Table \ref{fig:quant_div}, we show the error-of-best (i.e. pick the
colorization with minimum error to ground-truth) and the variance of
diverse colorizations. Lower error-of-best implies one of
the diverse predictions is close to ground-truth. Note that, our method reliably produces 
high variance with comparable error-of-best to other methods. Our goal is to generate diverse 
colorizations. However, since diverse colorizations are not observed in the ground-truth
for a single image, we cannot reliably evaluate them. Therefore, we use the weaker 
proxy of variance to evaluate diversity. Large variance is desirable for diverse 
colorization, which we obtain. We rely on qualitative evaluation to verify the 
naturalness of the different colorizations in the predicted pool.

\section{Conclusion}

Our loss terms help us build a variational autoencoder for high fidelity color fields.
The multi-modal conditional model produces embeddings that decode to realistic 
diverse colorizations. The colorizations obtained from our methods are more
diverse than CVAE and cGAN. The proposed method can be applied to other ambiguous 
problems. Our low dimensional embeddings allow us to predict diversity with multi-modal 
conditional models, but they do not encode high spatial detail. In future, our work will 
be focused on improving the spatial detail along with diversity. \\ 

\noindent
{\bf Acknowledgements.} We thank Arun Mallya and Jason Rock for useful 
discussions and suggestions. This work is supported in part by ONR MURI 
Award N00014-16-1-2007, and in part by NSF under Grants No.\ NSF 
IIS-1421521.

\clearpage
{\small
\bibliographystyle{ieee}
\bibliography{ref}
}

\appendix

\begin{center}
\begin{figure*}[!t]
\captionsetup[subfigure]{labelformat=empty}
\subfloat{\includegraphics[width=.08\textwidth]{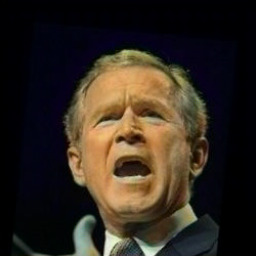}}  
\subfloat{\includegraphics[width=.08\textwidth]{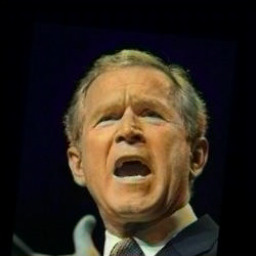}}  
\subfloat[cGAN]{\includegraphics[width=.08\textwidth]{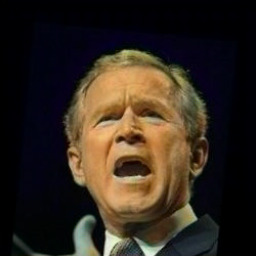}}  
\subfloat{\includegraphics[width=.08\textwidth]{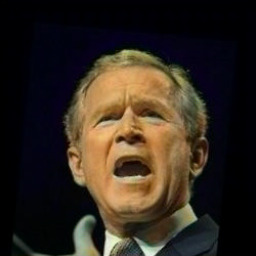}}  
\subfloat{\includegraphics[width=.08\textwidth]{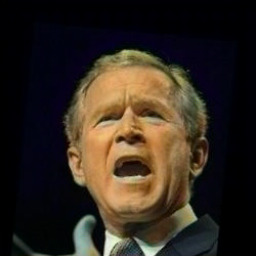}}  \hspace{8pt} 
\subfloat{\includegraphics[width=.08\textwidth]{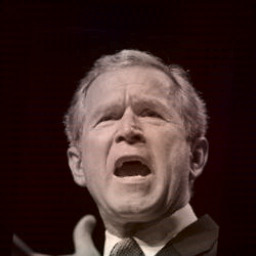}}  
\subfloat{\includegraphics[width=.08\textwidth]{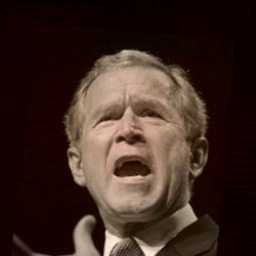}}  
\subfloat[CVAE]{\includegraphics[width=.08\textwidth]{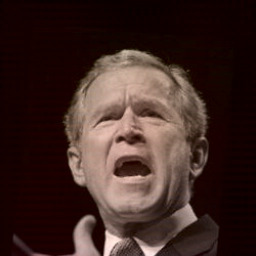}}  
\subfloat{\includegraphics[width=.08\textwidth]{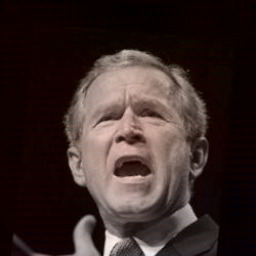}}  
\subfloat{\includegraphics[width=.08\textwidth]{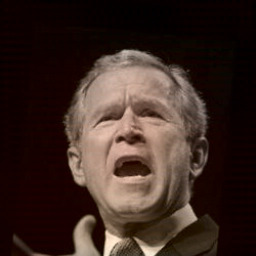}} \hspace{8pt}
\subfloat[GT]{\includegraphics[width=.08\textwidth]{lfw/gt/George_W_Bush_0503.jpg}} \\[-2.5ex]
\subfloat{\includegraphics[width=.08\textwidth]{lfw/cvpr_res/George_W_Bush_0503.jpg/divcolor_000.jpg}}  
\subfloat{\includegraphics[width=.08\textwidth]{lfw/cvpr_res/George_W_Bush_0503.jpg/divcolor_001.jpg}}  
\subfloat[Ours]{\includegraphics[width=.08\textwidth]{lfw/cvpr_res/George_W_Bush_0503.jpg/divcolor_002.jpg}}  
\subfloat{\includegraphics[width=.08\textwidth]{lfw/cvpr_res/George_W_Bush_0503.jpg/divcolor_003.jpg}}  
\subfloat{\includegraphics[width=.08\textwidth]{lfw/cvpr_res/George_W_Bush_0503.jpg/divcolor_004.jpg}}  \hspace{8pt} 
\subfloat{\includegraphics[width=.08\textwidth]{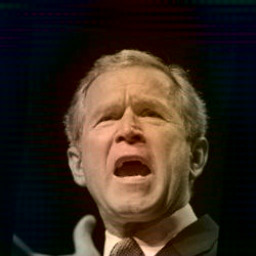}}  
\subfloat{\includegraphics[width=.08\textwidth]{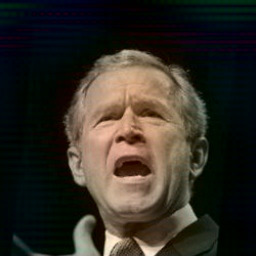}}  
\subfloat[Ours+Skip]{\includegraphics[width=.08\textwidth]{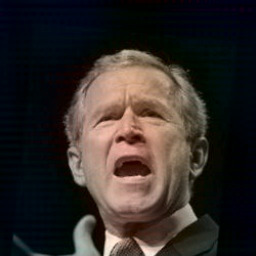}}  
\subfloat{\includegraphics[width=.08\textwidth]{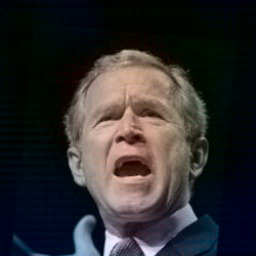}}  
\subfloat{\includegraphics[width=.08\textwidth]{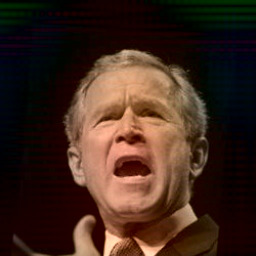}} \\ [-2ex]
\subfloat{\includegraphics[width=.08\textwidth]{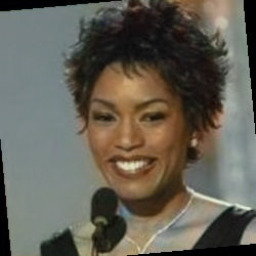}}  
\subfloat{\includegraphics[width=.08\textwidth]{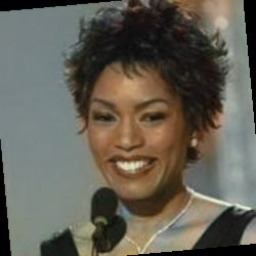}}  
\subfloat[cGAN]{\includegraphics[width=.08\textwidth]{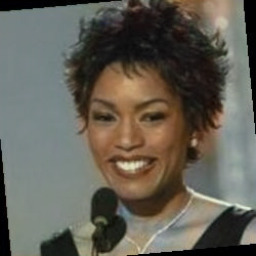}}  
\subfloat{\includegraphics[width=.08\textwidth]{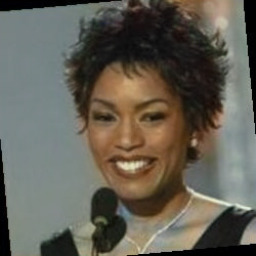}}  
\subfloat{\includegraphics[width=.08\textwidth]{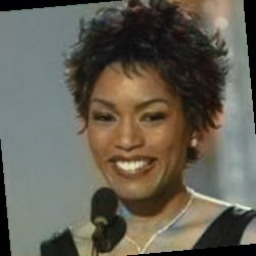}}  \hspace{8pt} 
\subfloat{\includegraphics[width=.08\textwidth]{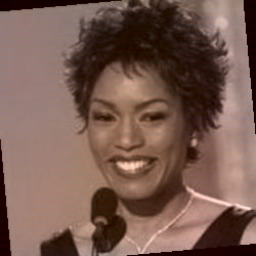}}  
\subfloat{\includegraphics[width=.08\textwidth]{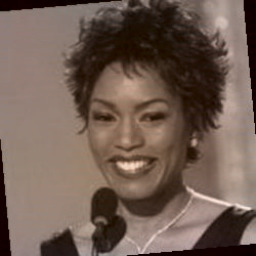}}  
\subfloat[CVAE]{\includegraphics[width=.08\textwidth]{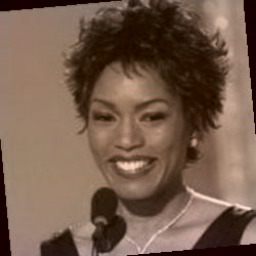}}  
\subfloat{\includegraphics[width=.08\textwidth]{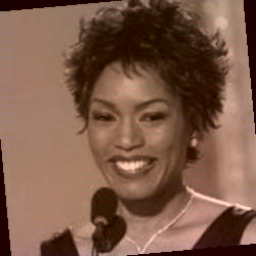}}  
\subfloat{\includegraphics[width=.08\textwidth]{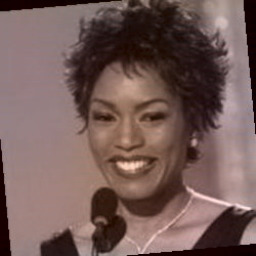}} \hspace{8pt}
\subfloat[GT]{\includegraphics[width=.08\textwidth]{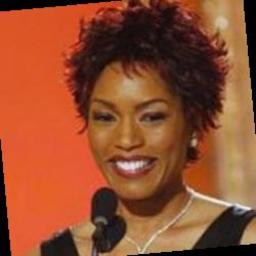}} \\ [-2.5ex]
\subfloat{\includegraphics[width=.08\textwidth]{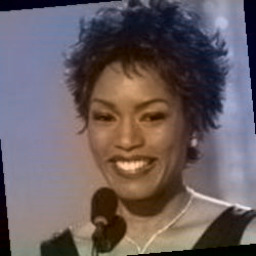}}  
\subfloat{\includegraphics[width=.08\textwidth]{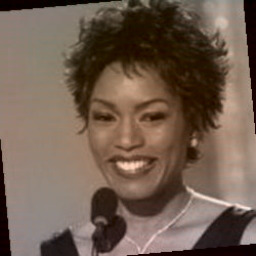}}  
\subfloat[Ours]{\includegraphics[width=.08\textwidth]{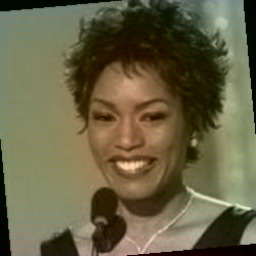}}  
\subfloat{\includegraphics[width=.08\textwidth]{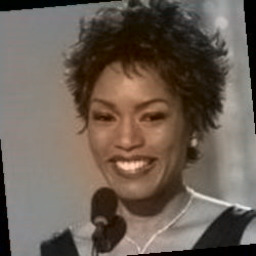}}  
\subfloat{\includegraphics[width=.08\textwidth]{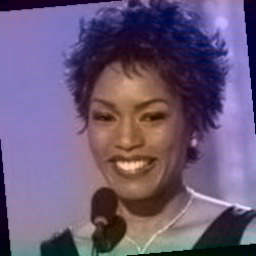}}  \hspace{8pt} 
\subfloat{\includegraphics[width=.08\textwidth]{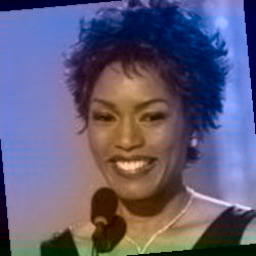}}  
\subfloat{\includegraphics[width=.08\textwidth]{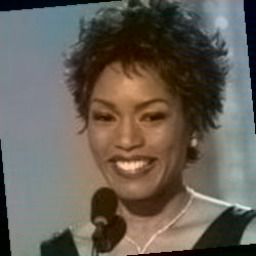}}  
\subfloat[Ours+Skip]{\includegraphics[width=.08\textwidth]{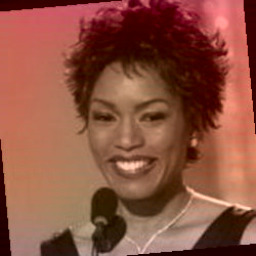}}  
\subfloat{\includegraphics[width=.08\textwidth]{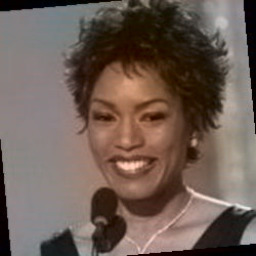}}  
\subfloat{\includegraphics[width=.08\textwidth]{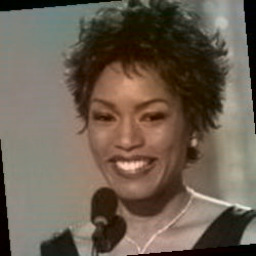}} \\[-2ex]
\subfloat{\includegraphics[width=.08\textwidth]{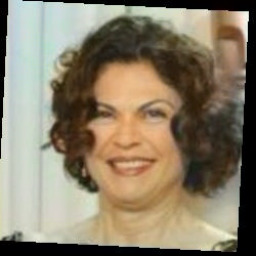}}  
\subfloat{\includegraphics[width=.08\textwidth]{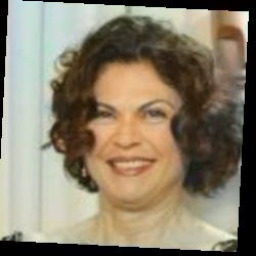}}  
\subfloat[cGAN]{\includegraphics[width=.08\textwidth]{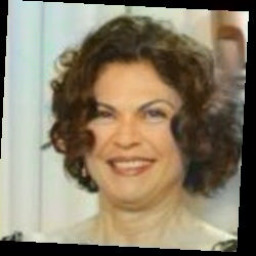}}  
\subfloat{\includegraphics[width=.08\textwidth]{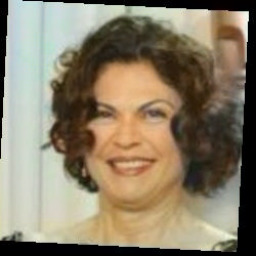}}  
\subfloat{\includegraphics[width=.08\textwidth]{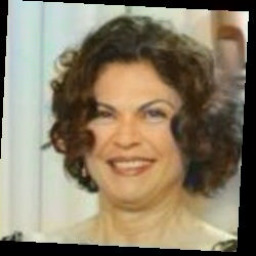}}  \hspace{8pt} 
\subfloat{\includegraphics[width=.08\textwidth]{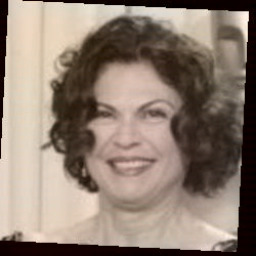}}  
\subfloat{\includegraphics[width=.08\textwidth]{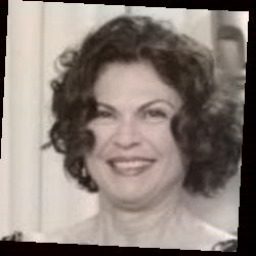}}  
\subfloat[CVAE]{\includegraphics[width=.08\textwidth]{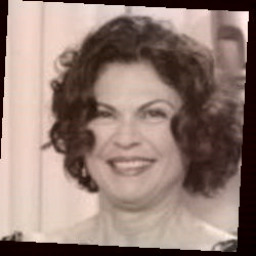}}  
\subfloat{\includegraphics[width=.08\textwidth]{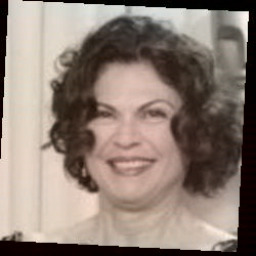}}  
\subfloat{\includegraphics[width=.08\textwidth]{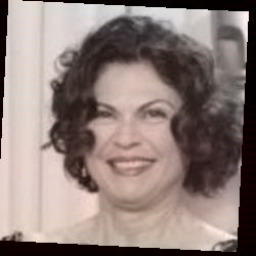}} \hspace{8pt}
\subfloat[GT]{\includegraphics[width=.08\textwidth]{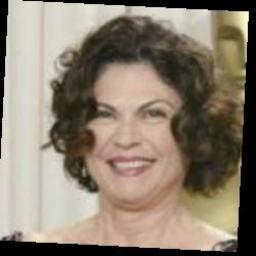}} \\[-2.5ex]
\subfloat{\includegraphics[width=.08\textwidth]{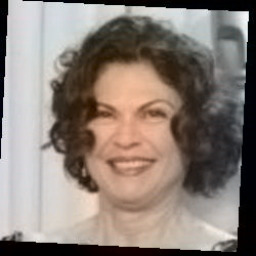}}  
\subfloat{\includegraphics[width=.08\textwidth]{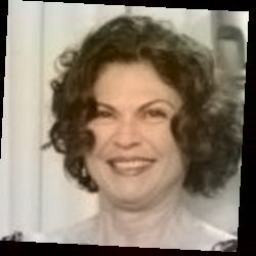}}  
\subfloat[Ours]{\includegraphics[width=.08\textwidth]{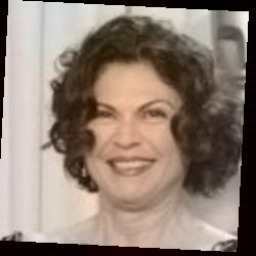}}  
\subfloat{\includegraphics[width=.08\textwidth]{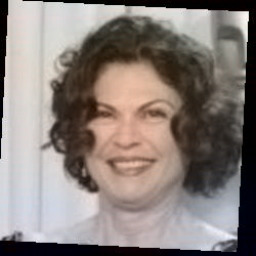}}  
\subfloat{\includegraphics[width=.08\textwidth]{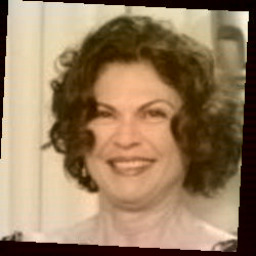}}  \hspace{8pt} 
\subfloat{\includegraphics[width=.08\textwidth]{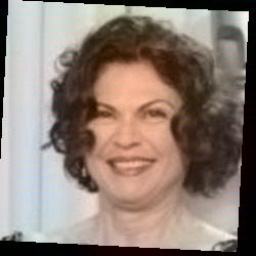}}  
\subfloat{\includegraphics[width=.08\textwidth]{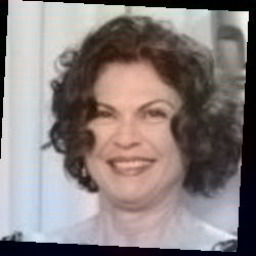}}  
\subfloat[Ours+Skip]{\includegraphics[width=.08\textwidth]{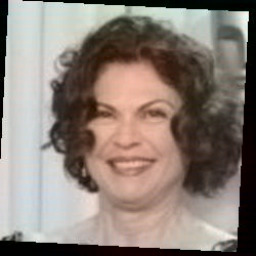}}  
\subfloat{\includegraphics[width=.08\textwidth]{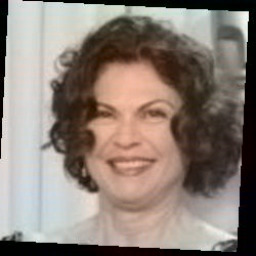}}  
\subfloat{\includegraphics[width=.08\textwidth]{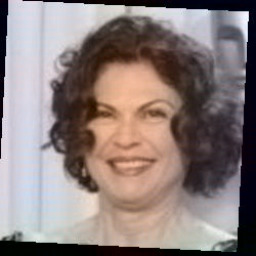}} \\ [-2ex]
\subfloat{\includegraphics[width=.08\textwidth]{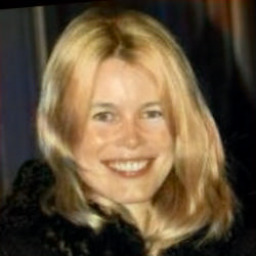}}  
\subfloat{\includegraphics[width=.08\textwidth]{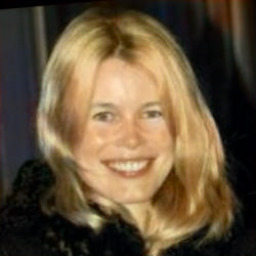}}  
\subfloat[cGAN]{\includegraphics[width=.08\textwidth]{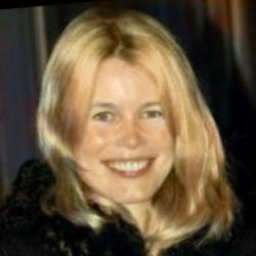}}  
\subfloat{\includegraphics[width=.08\textwidth]{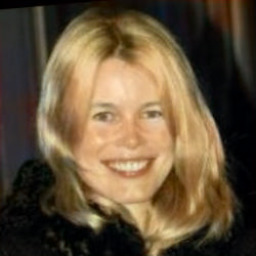}}  
\subfloat{\includegraphics[width=.08\textwidth]{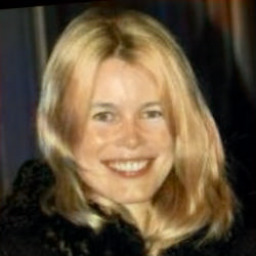}}  \hspace{8pt} 
\subfloat{\includegraphics[width=.08\textwidth]{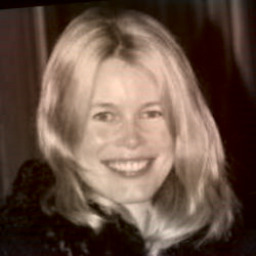}}  
\subfloat{\includegraphics[width=.08\textwidth]{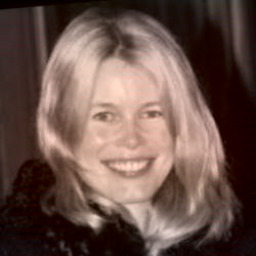}}  
\subfloat[CVAE]{\includegraphics[width=.08\textwidth]{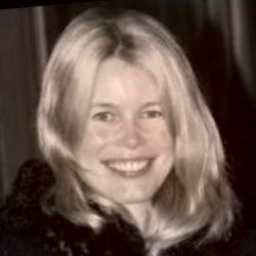}}  
\subfloat{\includegraphics[width=.08\textwidth]{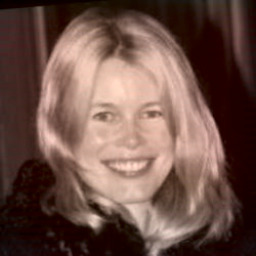}}  
\subfloat{\includegraphics[width=.08\textwidth]{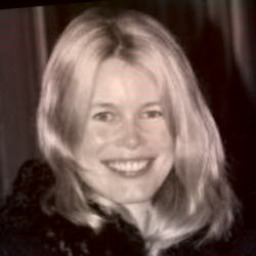}} \hspace{8pt}
\subfloat[GT]{\includegraphics[width=.08\textwidth]{lfw/gt/Claudia_Schiffer_0003.jpg}} \\ [-2.5ex]
\subfloat{\includegraphics[width=.08\textwidth]{lfw/cvpr_res/Claudia_Schiffer_0003.jpg/divcolor_000.jpg}}  
\subfloat{\includegraphics[width=.08\textwidth]{lfw/cvpr_res/Claudia_Schiffer_0003.jpg/divcolor_001.jpg}}  
\subfloat[Ours]{\includegraphics[width=.08\textwidth]{lfw/cvpr_res/Claudia_Schiffer_0003.jpg/divcolor_002.jpg}}  
\subfloat{\includegraphics[width=.08\textwidth]{lfw/cvpr_res/Claudia_Schiffer_0003.jpg/divcolor_003.jpg}}  
\subfloat{\includegraphics[width=.08\textwidth]{lfw/cvpr_res/Claudia_Schiffer_0003.jpg/divcolor_004.jpg}}  \hspace{8pt} 
\subfloat{\includegraphics[width=.08\textwidth]{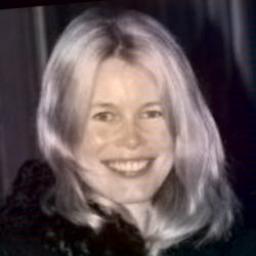}}  
\subfloat{\includegraphics[width=.08\textwidth]{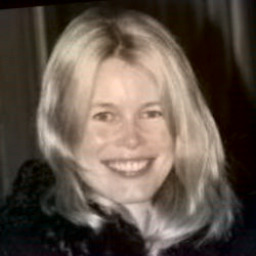}}  
\subfloat[Ours+Skip]{\includegraphics[width=.08\textwidth]{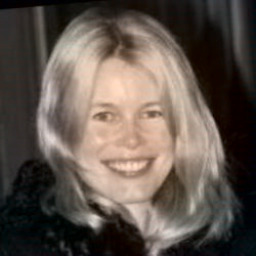}}  
\subfloat{\includegraphics[width=.08\textwidth]{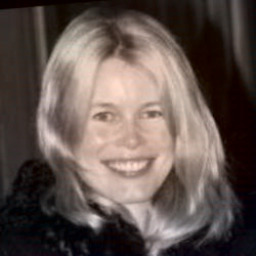}}  
\subfloat{\includegraphics[width=.08\textwidth]{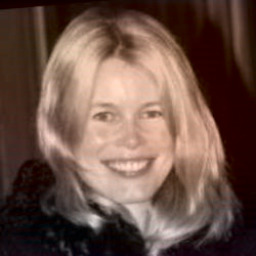}} \\[-2ex]
\subfloat{\includegraphics[width=.08\textwidth]{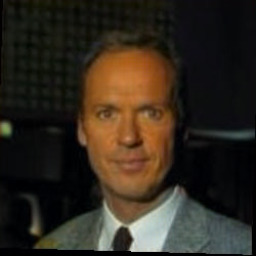}}  
\subfloat{\includegraphics[width=.08\textwidth]{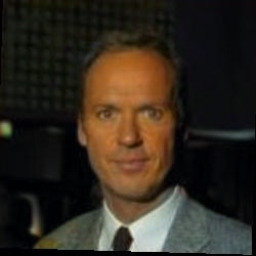}}  
\subfloat[cGAN]{\includegraphics[width=.08\textwidth]{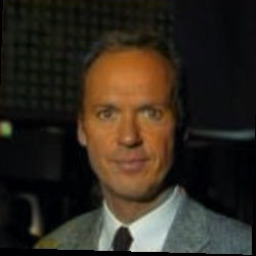}}  
\subfloat{\includegraphics[width=.08\textwidth]{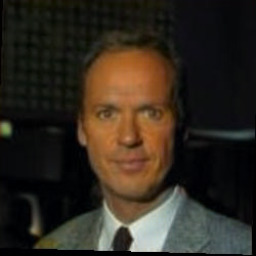}}  
\subfloat{\includegraphics[width=.08\textwidth]{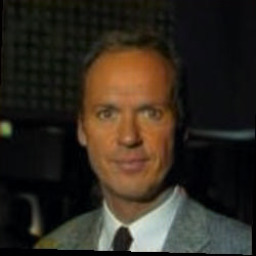}}  \hspace{8pt} 
\subfloat{\includegraphics[width=.08\textwidth]{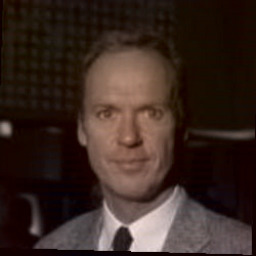}}  
\subfloat{\includegraphics[width=.08\textwidth]{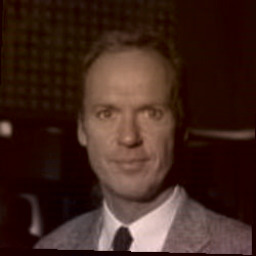}}  
\subfloat[CVAE]{\includegraphics[width=.08\textwidth]{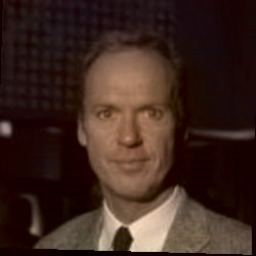}}  
\subfloat{\includegraphics[width=.08\textwidth]{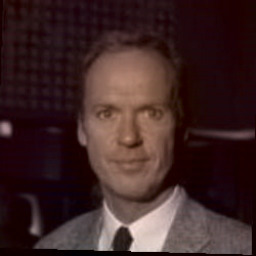}}  
\subfloat{\includegraphics[width=.08\textwidth]{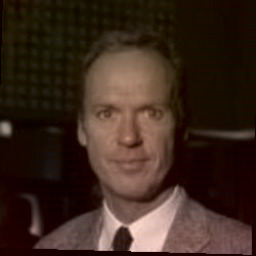}} \hspace{8pt}
\subfloat[GT]{\includegraphics[width=.08\textwidth]{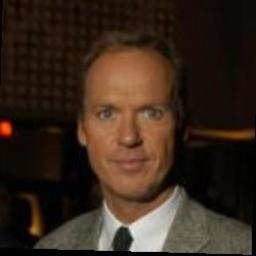}} \\[-2.5ex]
\subfloat{\includegraphics[width=.08\textwidth]{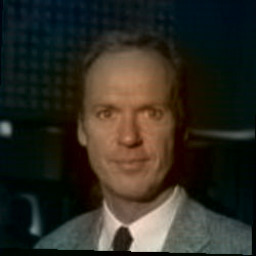}}  
\subfloat{\includegraphics[width=.08\textwidth]{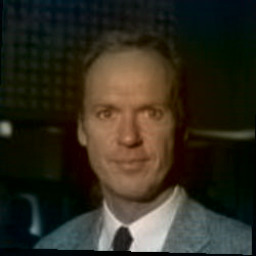}}  
\subfloat[Ours]{\includegraphics[width=.08\textwidth]{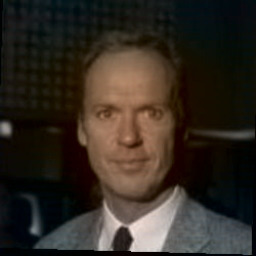}}  
\subfloat{\includegraphics[width=.08\textwidth]{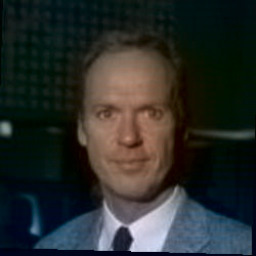}}  
\subfloat{\includegraphics[width=.08\textwidth]{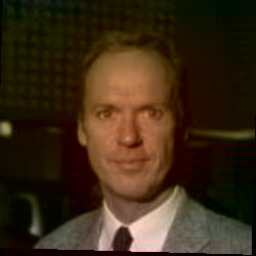}}  \hspace{8pt} 
\subfloat{\includegraphics[width=.08\textwidth]{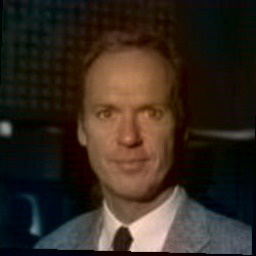}}  
\subfloat{\includegraphics[width=.08\textwidth]{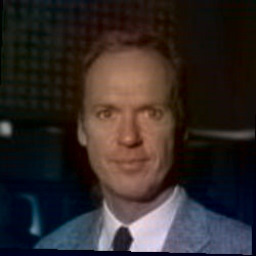}}  
\subfloat[Ours+Skip]{\includegraphics[width=.08\textwidth]{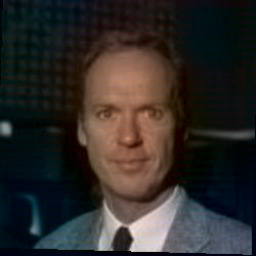}}  
\subfloat{\includegraphics[width=.08\textwidth]{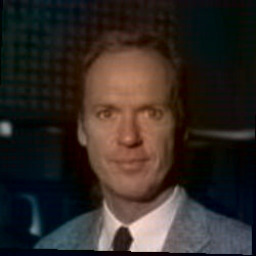}}  
\subfloat{\includegraphics[width=.08\textwidth]{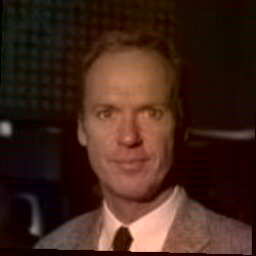}} \\ [-2ex]

\caption{For LFW dataset, diverse colorizations from our methods are compared to the CVAE, cGAN and the ground-truth (GT).} 
\label{fig:res_lfw} 
\end{figure*}
\end{center}

\begin{center}
\begin{figure*}[!t]
\captionsetup[subfigure]{labelformat=empty}
\subfloat{\includegraphics[width=.08\textwidth]{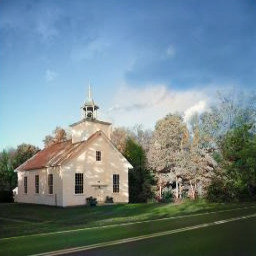}}  
\subfloat{\includegraphics[width=.08\textwidth]{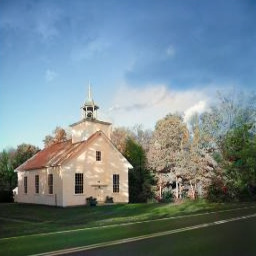}}  
\subfloat[cGAN]{\includegraphics[width=.08\textwidth]{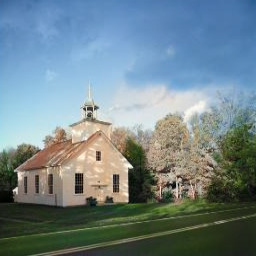}}  
\subfloat{\includegraphics[width=.08\textwidth]{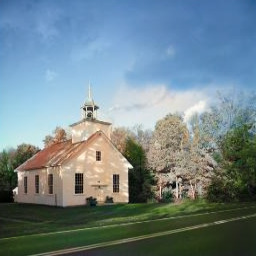}}  
\subfloat{\includegraphics[width=.08\textwidth]{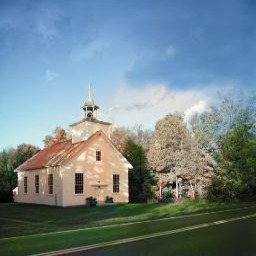}}  \hspace{8pt} 
\subfloat{\includegraphics[width=.08\textwidth]{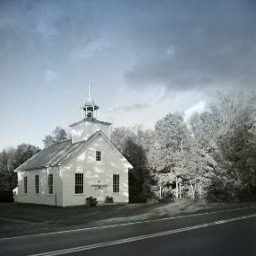}}  
\subfloat{\includegraphics[width=.08\textwidth]{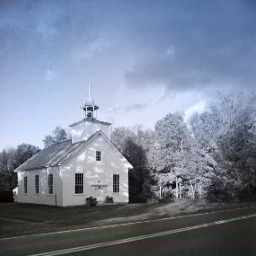}}  
\subfloat[CVAE]{\includegraphics[width=.08\textwidth]{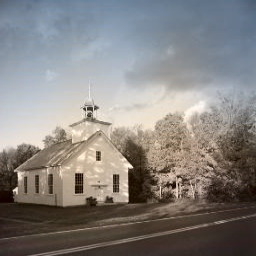}}  
\subfloat{\includegraphics[width=.08\textwidth]{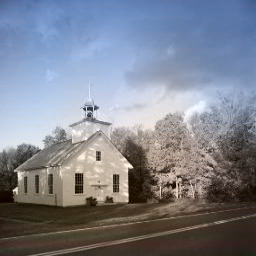}}  
\subfloat{\includegraphics[width=.08\textwidth]{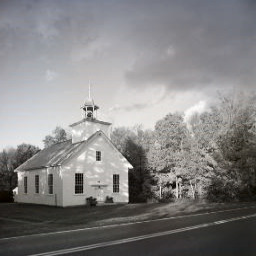}} \hspace{8pt}
\subfloat[GT]{\includegraphics[width=.08\textwidth]{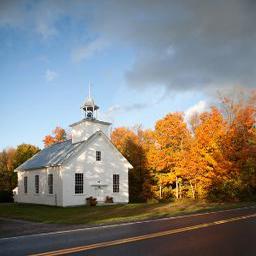}} \\[-2.5ex]
\subfloat{\includegraphics[width=.08\textwidth]{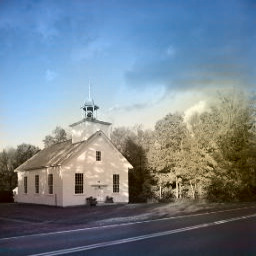}}  
\subfloat{\includegraphics[width=.08\textwidth]{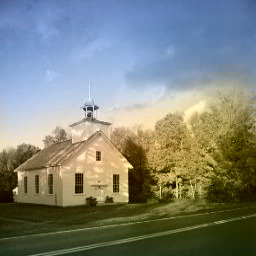}}  
\subfloat[Ours]{\includegraphics[width=.08\textwidth]{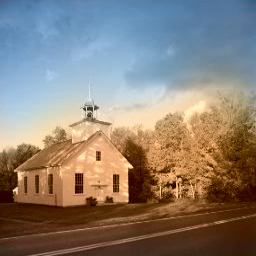}}  
\subfloat{\includegraphics[width=.08\textwidth]{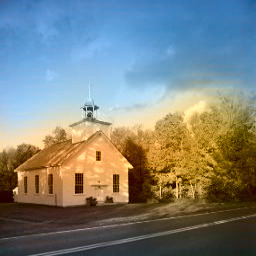}}  
\subfloat{\includegraphics[width=.08\textwidth]{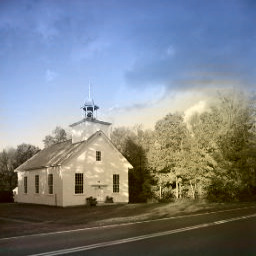}}  \hspace{8pt} 
\subfloat{\includegraphics[width=.08\textwidth]{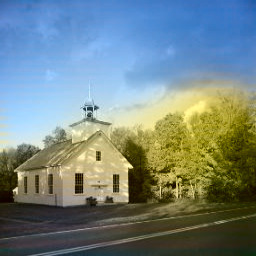}}  
\subfloat{\includegraphics[width=.08\textwidth]{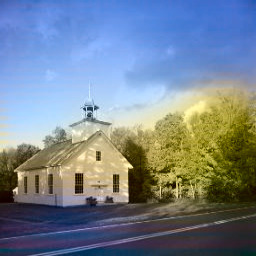}}  
\subfloat[Ours+Skip]{\includegraphics[width=.08\textwidth]{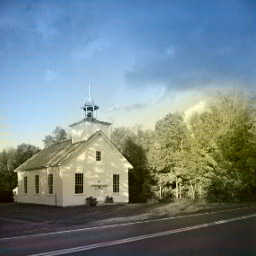}}  
\subfloat{\includegraphics[width=.08\textwidth]{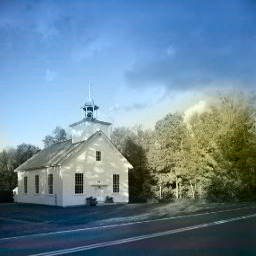}}  
\subfloat{\includegraphics[width=.08\textwidth]{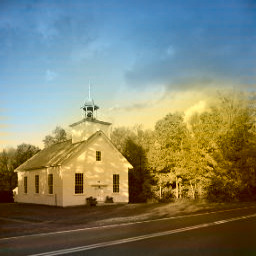}} \\[-2ex]
\subfloat{\includegraphics[width=.08\textwidth]{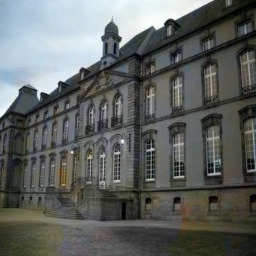}}  
\subfloat{\includegraphics[width=.08\textwidth]{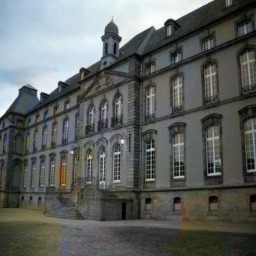}}  
\subfloat[cGAN]{\includegraphics[width=.08\textwidth]{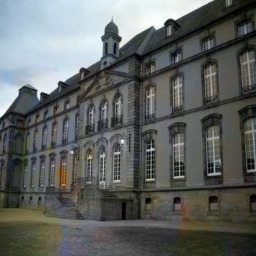}}  
\subfloat{\includegraphics[width=.08\textwidth]{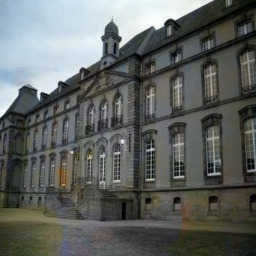}}  
\subfloat{\includegraphics[width=.08\textwidth]{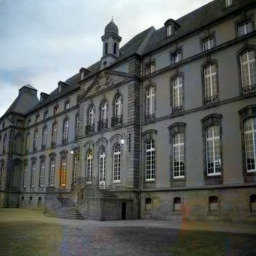}}  \hspace{8pt} 
\subfloat{\includegraphics[width=.08\textwidth]{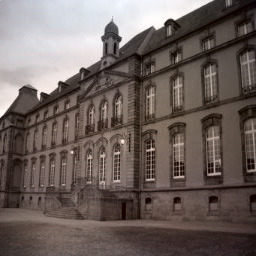}}  
\subfloat{\includegraphics[width=.08\textwidth]{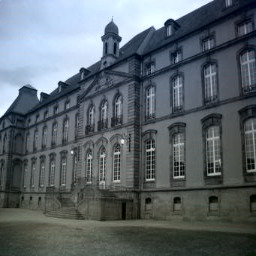}}  
\subfloat[CVAE]{\includegraphics[width=.08\textwidth]{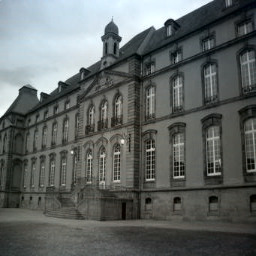}}  
\subfloat{\includegraphics[width=.08\textwidth]{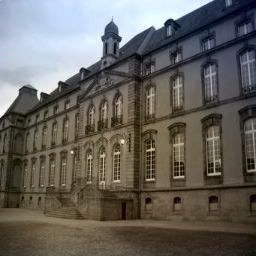}}  
\subfloat{\includegraphics[width=.08\textwidth]{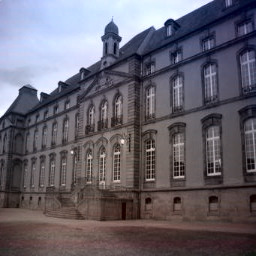}} \hspace{8pt}
\subfloat[GT]{\includegraphics[width=.08\textwidth]{church/gt/01a392445923d690614224cd3a4d4d6d1c7fa2bd.jpg}} \\[-2.5ex]
\subfloat{\includegraphics[width=.08\textwidth]{church/cvpr_res/01a392445923d690614224cd3a4d4d6d1c7fa2bd.webp/divcolor_000.jpg}}  
\subfloat{\includegraphics[width=.08\textwidth]{church/cvpr_res/01a392445923d690614224cd3a4d4d6d1c7fa2bd.webp/divcolor_001.jpg}}  
\subfloat[Ours]{\includegraphics[width=.08\textwidth]{church/cvpr_res/01a392445923d690614224cd3a4d4d6d1c7fa2bd.webp/divcolor_002.jpg}}  
\subfloat{\includegraphics[width=.08\textwidth]{church/cvpr_res/01a392445923d690614224cd3a4d4d6d1c7fa2bd.webp/divcolor_003.jpg}}  
\subfloat{\includegraphics[width=.08\textwidth]{church/cvpr_res/01a392445923d690614224cd3a4d4d6d1c7fa2bd.webp/divcolor_004.jpg}}  \hspace{8pt} 
\subfloat{\includegraphics[width=.08\textwidth]{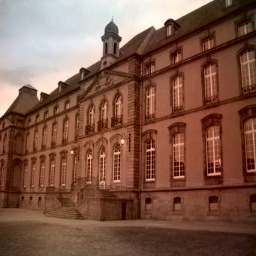}}  
\subfloat{\includegraphics[width=.08\textwidth]{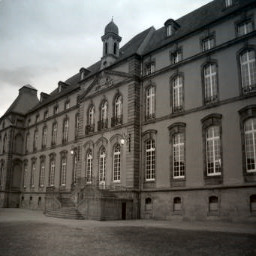}}  
\subfloat[Ours+Skip]{\includegraphics[width=.08\textwidth]{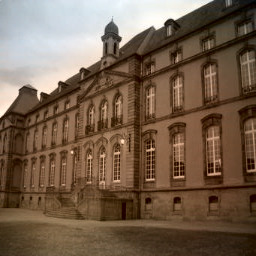}}  
\subfloat{\includegraphics[width=.08\textwidth]{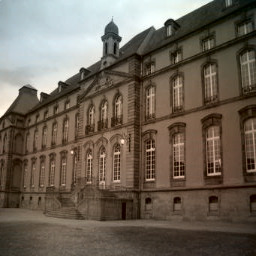}}  
\subfloat{\includegraphics[width=.08\textwidth]{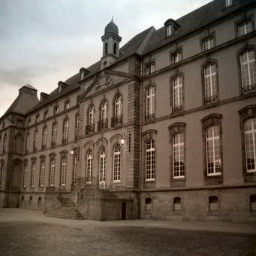}} \\ [-2ex]
\subfloat{\includegraphics[width=.08\textwidth]{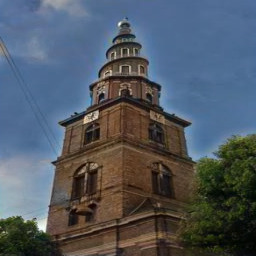}}  
\subfloat{\includegraphics[width=.08\textwidth]{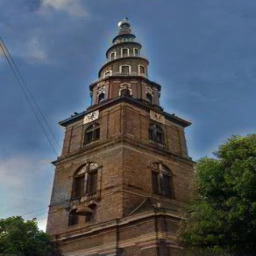}}  
\subfloat[cGAN]{\includegraphics[width=.08\textwidth]{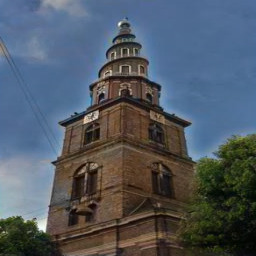}}  
\subfloat{\includegraphics[width=.08\textwidth]{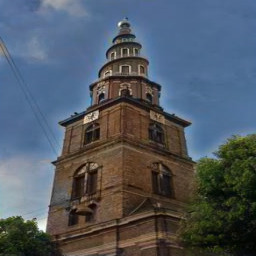}}  
\subfloat{\includegraphics[width=.08\textwidth]{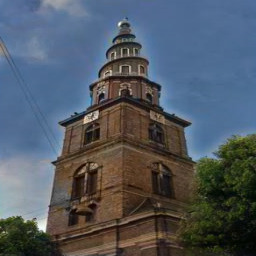}}  \hspace{8pt} 
\subfloat{\includegraphics[width=.08\textwidth]{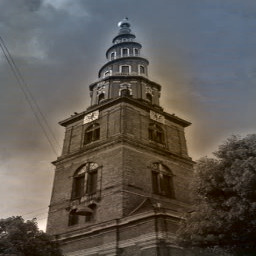}}  
\subfloat{\includegraphics[width=.08\textwidth]{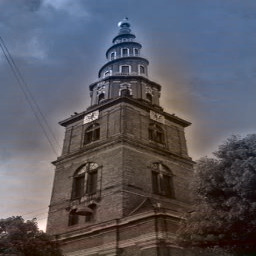}}  
\subfloat[CVAE]{\includegraphics[width=.08\textwidth]{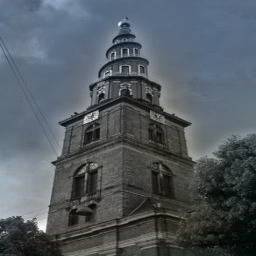}}  
\subfloat{\includegraphics[width=.08\textwidth]{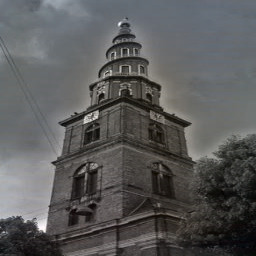}}  
\subfloat{\includegraphics[width=.08\textwidth]{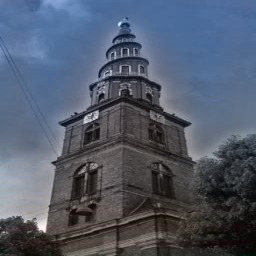}} \hspace{8pt}
\subfloat[GT]{\includegraphics[width=.08\textwidth]{church/gt/7145fdad2e4dcd0d67e3f5b9c419a7d14bc5e84d.jpg}} \\[-2.5ex]
\subfloat{\includegraphics[width=.08\textwidth]{church/cvpr_res/7145fdad2e4dcd0d67e3f5b9c419a7d14bc5e84d.webp/divcolor_000.jpg}}  
\subfloat{\includegraphics[width=.08\textwidth]{church/cvpr_res/7145fdad2e4dcd0d67e3f5b9c419a7d14bc5e84d.webp/divcolor_001.jpg}}  
\subfloat[Ours]{\includegraphics[width=.08\textwidth]{church/cvpr_res/7145fdad2e4dcd0d67e3f5b9c419a7d14bc5e84d.webp/divcolor_002.jpg}}  
\subfloat{\includegraphics[width=.08\textwidth]{church/cvpr_res/7145fdad2e4dcd0d67e3f5b9c419a7d14bc5e84d.webp/divcolor_003.jpg}}  
\subfloat{\includegraphics[width=.08\textwidth]{church/cvpr_res/7145fdad2e4dcd0d67e3f5b9c419a7d14bc5e84d.webp/divcolor_004.jpg}}  \hspace{8pt} 
\subfloat{\includegraphics[width=.08\textwidth]{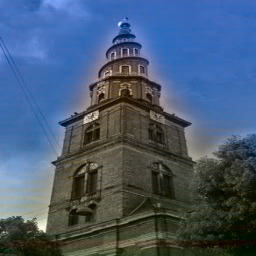}}  
\subfloat{\includegraphics[width=.08\textwidth]{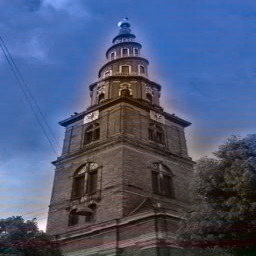}}  
\subfloat[Ours+Skip]{\includegraphics[width=.08\textwidth]{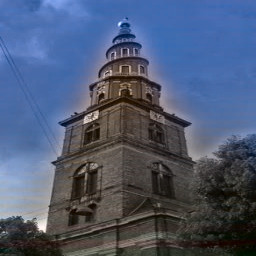}}  
\subfloat{\includegraphics[width=.08\textwidth]{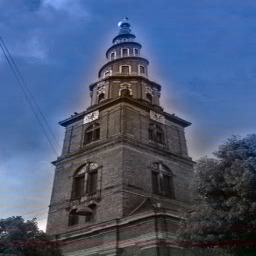}}  
\subfloat{\includegraphics[width=.08\textwidth]{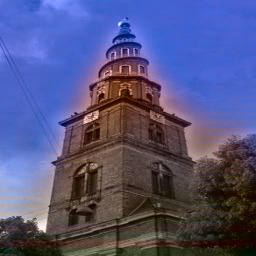}} \\[-2ex]
\subfloat{\includegraphics[width=.08\textwidth]{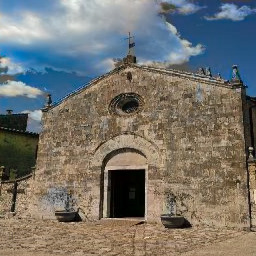}}  
\subfloat{\includegraphics[width=.08\textwidth]{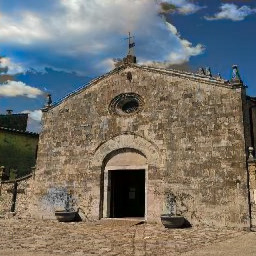}}  
\subfloat[cGAN]{\includegraphics[width=.08\textwidth]{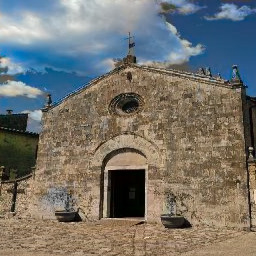}}  
\subfloat{\includegraphics[width=.08\textwidth]{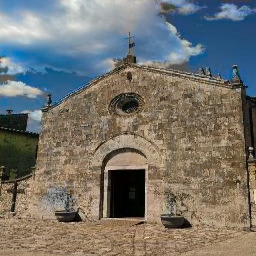}}  
\subfloat{\includegraphics[width=.08\textwidth]{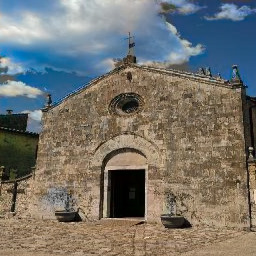}}  \hspace{8pt} 
\subfloat{\includegraphics[width=.08\textwidth]{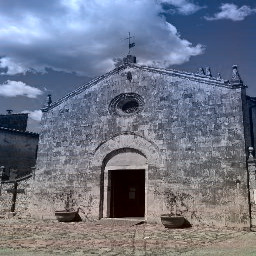}}  
\subfloat{\includegraphics[width=.08\textwidth]{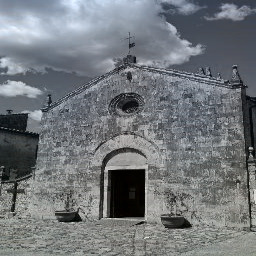}}  
\subfloat[CVAE]{\includegraphics[width=.08\textwidth]{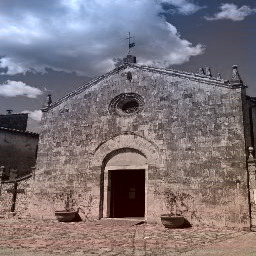}}  
\subfloat{\includegraphics[width=.08\textwidth]{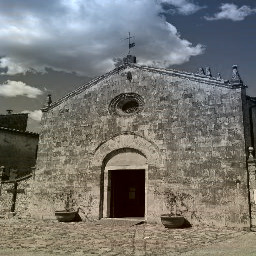}}  
\subfloat{\includegraphics[width=.08\textwidth]{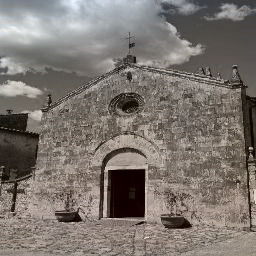}} \hspace{8pt}
\subfloat[GT]{\includegraphics[width=.08\textwidth]{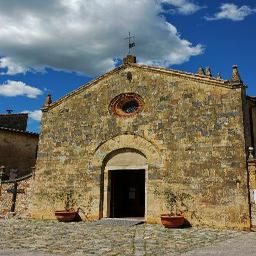}} \\[-2.5ex]
\subfloat{\includegraphics[width=.08\textwidth]{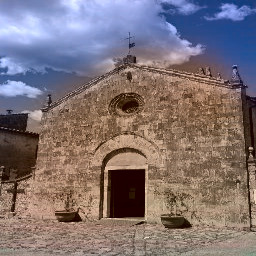}}  
\subfloat{\includegraphics[width=.08\textwidth]{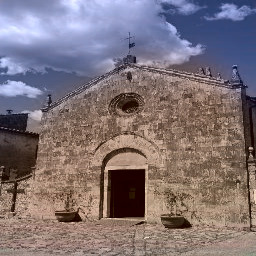}}  
\subfloat[Ours]{\includegraphics[width=.08\textwidth]{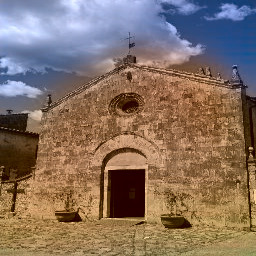}}  
\subfloat{\includegraphics[width=.08\textwidth]{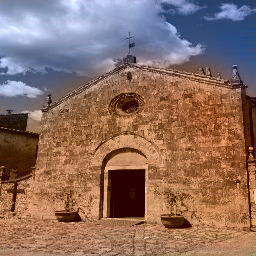}}  
\subfloat{\includegraphics[width=.08\textwidth]{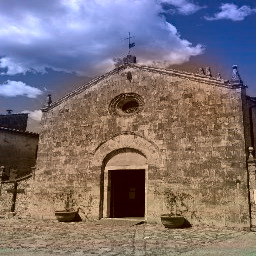}}  \hspace{8pt} 
\subfloat{\includegraphics[width=.08\textwidth]{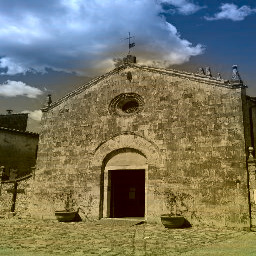}}  
\subfloat{\includegraphics[width=.08\textwidth]{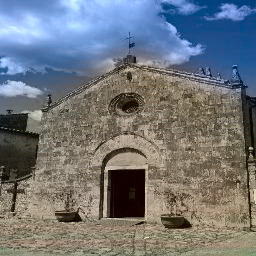}}  
\subfloat[Ours+Skip]{\includegraphics[width=.08\textwidth]{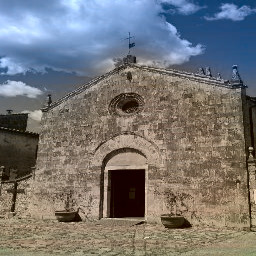}}  
\subfloat{\includegraphics[width=.08\textwidth]{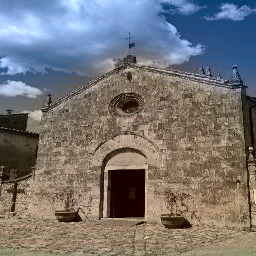}}  
\subfloat{\includegraphics[width=.08\textwidth]{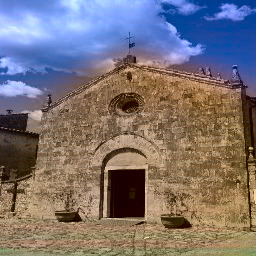}} \\ [-2ex]
\subfloat{\includegraphics[width=.08\textwidth]{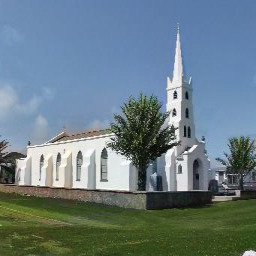}}  
\subfloat{\includegraphics[width=.08\textwidth]{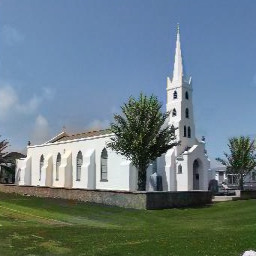}}  
\subfloat[cGAN]{\includegraphics[width=.08\textwidth]{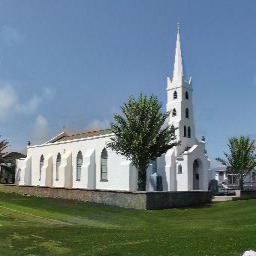}}  
\subfloat{\includegraphics[width=.08\textwidth]{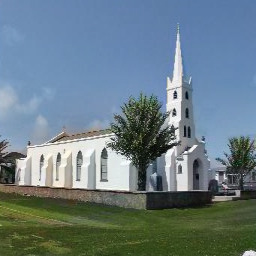}}  
\subfloat{\includegraphics[width=.08\textwidth]{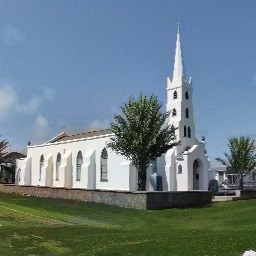}}  \hspace{8pt} 
\subfloat{\includegraphics[width=.08\textwidth]{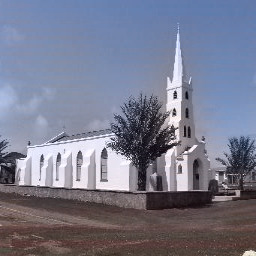}}  
\subfloat{\includegraphics[width=.08\textwidth]{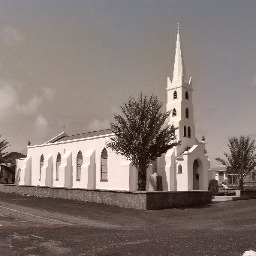}}  
\subfloat[CVAE]{\includegraphics[width=.08\textwidth]{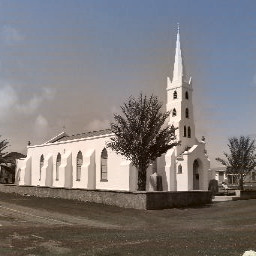}}  
\subfloat{\includegraphics[width=.08\textwidth]{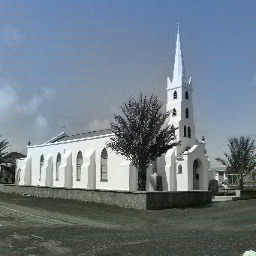}}  
\subfloat{\includegraphics[width=.08\textwidth]{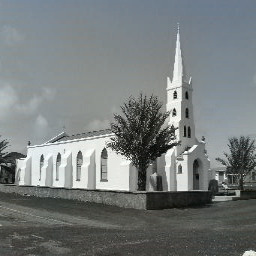}} \hspace{8pt}
\subfloat[GT]{\includegraphics[width=.08\textwidth]{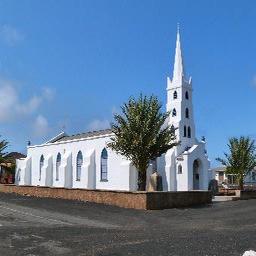}} \\[-2.5ex]
\subfloat{\includegraphics[width=.08\textwidth]{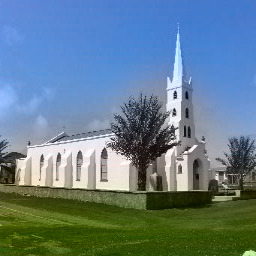}}  
\subfloat{\includegraphics[width=.08\textwidth]{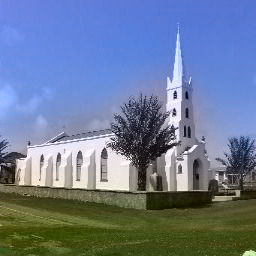}}  
\subfloat[Ours]{\includegraphics[width=.08\textwidth]{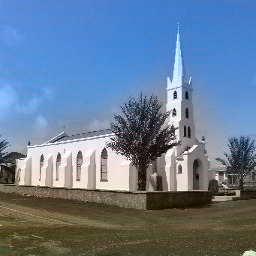}}  
\subfloat{\includegraphics[width=.08\textwidth]{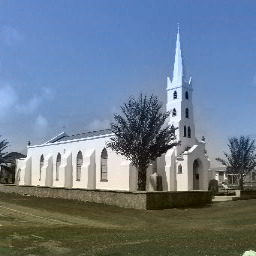}}  
\subfloat{\includegraphics[width=.08\textwidth]{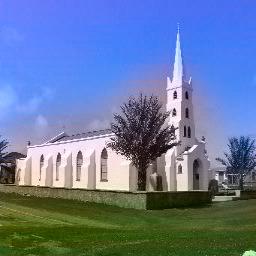}}  \hspace{8pt} 
\subfloat{\includegraphics[width=.08\textwidth]{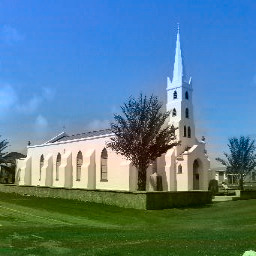}}  
\subfloat{\includegraphics[width=.08\textwidth]{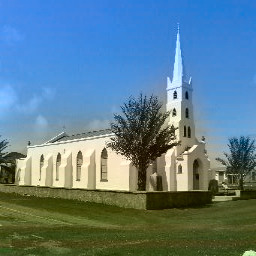}}  
\subfloat[Ours+Skip]{\includegraphics[width=.08\textwidth]{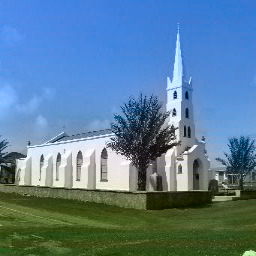}}  
\subfloat{\includegraphics[width=.08\textwidth]{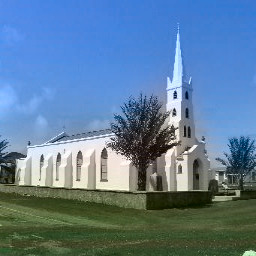}}  
\subfloat{\includegraphics[width=.08\textwidth]{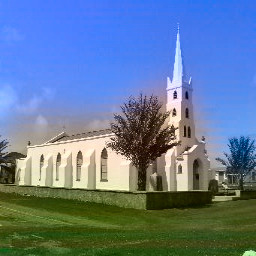}} \\[-2ex]
\caption{For LSUN Church dataset, diverse colorizations from our methods are compared to the CVAE, cGAN and the ground-truth (GT).} 
\label{fig:res_church} 
\end{figure*}
\end{center}

\begin{center}
\begin{figure*}[!t]
\captionsetup[subfigure]{labelformat=empty}
\subfloat{\includegraphics[width=.08\textwidth]{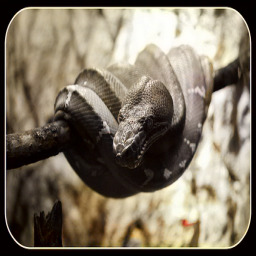}}  
\subfloat{\includegraphics[width=.08\textwidth]{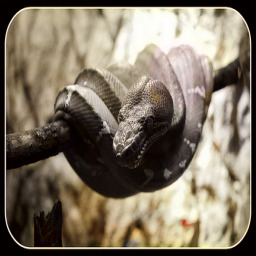}}  
\subfloat[cGAN]{\includegraphics[width=.08\textwidth]{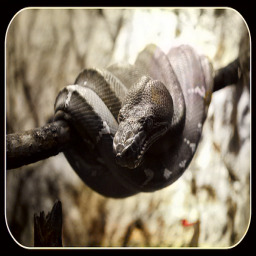}}  
\subfloat{\includegraphics[width=.08\textwidth]{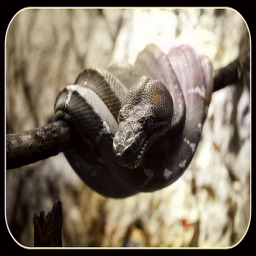}}  
\subfloat{\includegraphics[width=.08\textwidth]{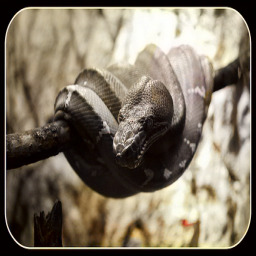}}  \hspace{8pt} 
\subfloat{\includegraphics[width=.08\textwidth]{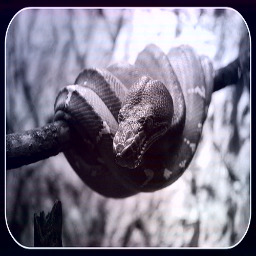}}  
\subfloat{\includegraphics[width=.08\textwidth]{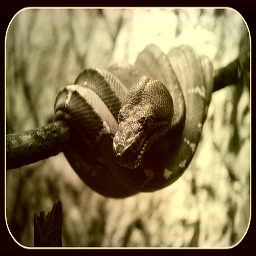}}  
\subfloat[CVAE]{\includegraphics[width=.08\textwidth]{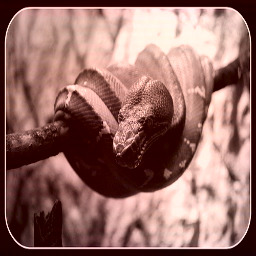}}  
\subfloat{\includegraphics[width=.08\textwidth]{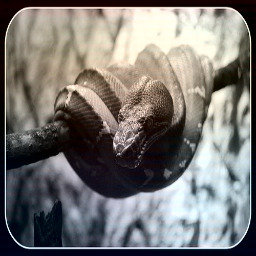}}  
\subfloat{\includegraphics[width=.08\textwidth]{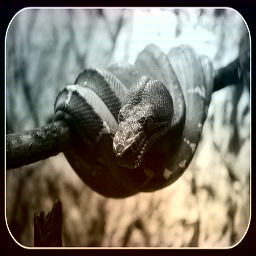}} \hspace{8pt}
\subfloat[GT]{\includegraphics[width=.08\textwidth]{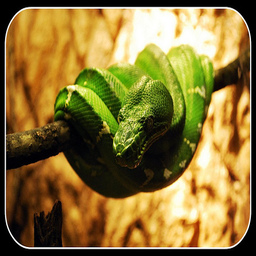}} \\[-2.5ex]
\subfloat{\includegraphics[width=.08\textwidth]{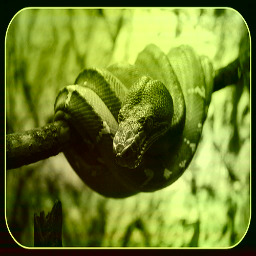}}  
\subfloat{\includegraphics[width=.08\textwidth]{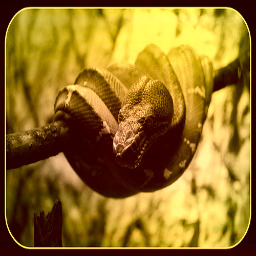}}  
\subfloat[Ours]{\includegraphics[width=.08\textwidth]{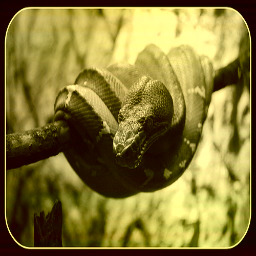}}  
\subfloat{\includegraphics[width=.08\textwidth]{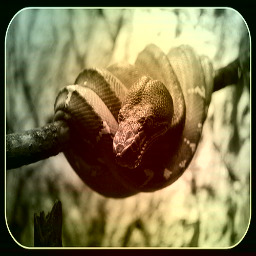}}  
\subfloat{\includegraphics[width=.08\textwidth]{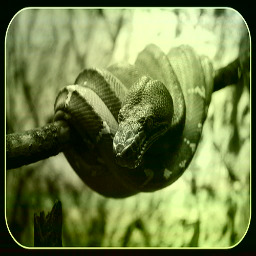}}  \hspace{8pt} 
\subfloat{\includegraphics[width=.08\textwidth]{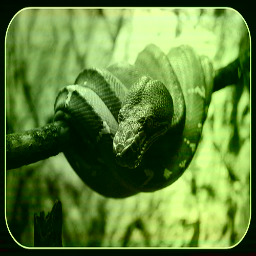}}  
\subfloat{\includegraphics[width=.08\textwidth]{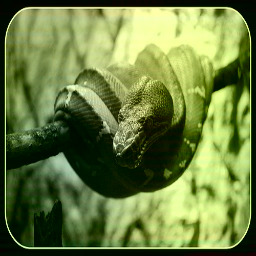}}  
\subfloat[Ours+Skip]{\includegraphics[width=.08\textwidth]{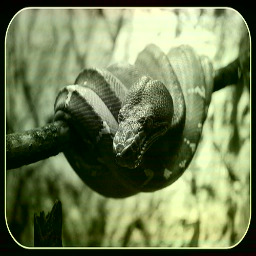}}  
\subfloat{\includegraphics[width=.08\textwidth]{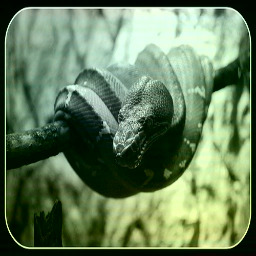}}  
\subfloat{\includegraphics[width=.08\textwidth]{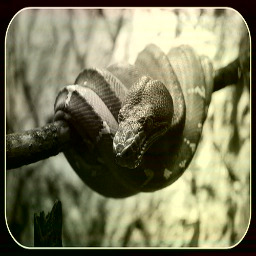}} \\ [-2ex]
\subfloat{\includegraphics[width=.08\textwidth]{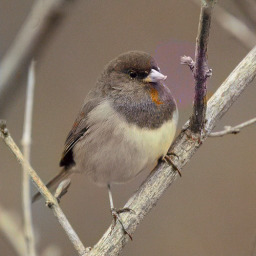}}  
\subfloat{\includegraphics[width=.08\textwidth]{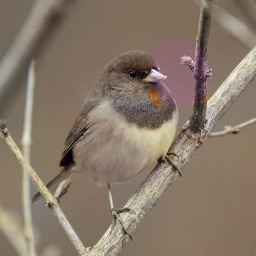}}  
\subfloat[cGAN]{\includegraphics[width=.08\textwidth]{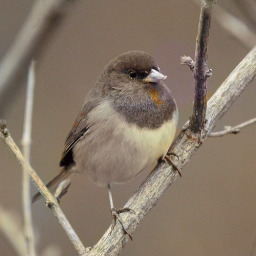}}  
\subfloat{\includegraphics[width=.08\textwidth]{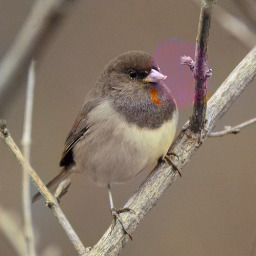}}  
\subfloat{\includegraphics[width=.08\textwidth]{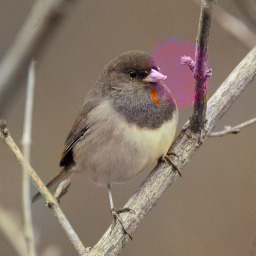}}  \hspace{8pt} 
\subfloat{\includegraphics[width=.08\textwidth]{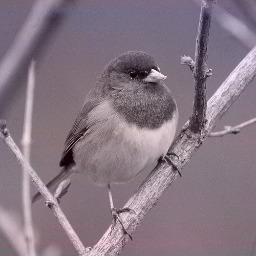}}  
\subfloat{\includegraphics[width=.08\textwidth]{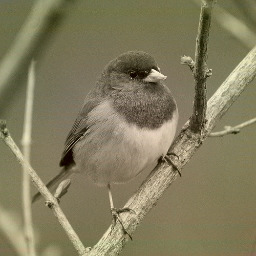}}  
\subfloat[CVAE]{\includegraphics[width=.08\textwidth]{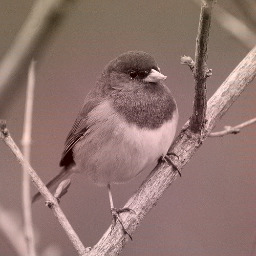}}  
\subfloat{\includegraphics[width=.08\textwidth]{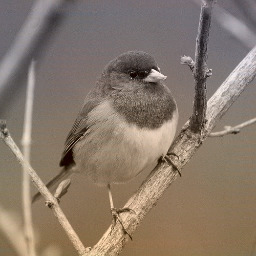}}  
\subfloat{\includegraphics[width=.08\textwidth]{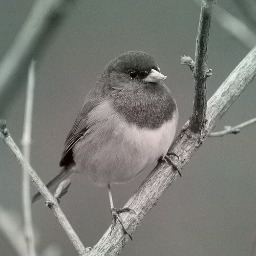}} \hspace{8pt}
\subfloat[GT]{\includegraphics[width=.08\textwidth]{imagenetval/gt/ILSVRC2012_val_00000247.jpg}} \\[-2.5ex]
\subfloat{\includegraphics[width=.08\textwidth]{imagenetval/cvpr_res/ILSVRC2012_val_00000247.JPEG/divcolor_000.jpg}}  
\subfloat{\includegraphics[width=.08\textwidth]{imagenetval/cvpr_res/ILSVRC2012_val_00000247.JPEG/divcolor_001.jpg}}  
\subfloat[Ours]{\includegraphics[width=.08\textwidth]{imagenetval/cvpr_res/ILSVRC2012_val_00000247.JPEG/divcolor_002.jpg}}  
\subfloat{\includegraphics[width=.08\textwidth]{imagenetval/cvpr_res/ILSVRC2012_val_00000247.JPEG/divcolor_003.jpg}}  
\subfloat{\includegraphics[width=.08\textwidth]{imagenetval/cvpr_res/ILSVRC2012_val_00000247.JPEG/divcolor_004.jpg}}  \hspace{8pt} 
\subfloat{\includegraphics[width=.08\textwidth]{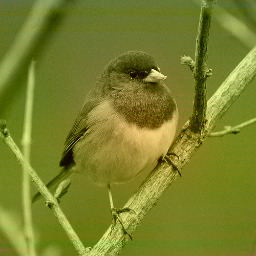}}  
\subfloat{\includegraphics[width=.08\textwidth]{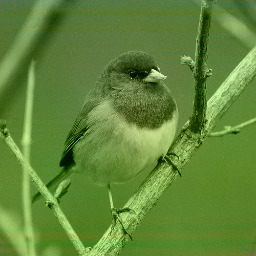}}  
\subfloat[Ours+Skip]{\includegraphics[width=.08\textwidth]{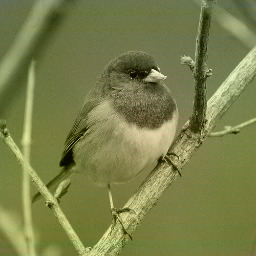}}  
\subfloat{\includegraphics[width=.08\textwidth]{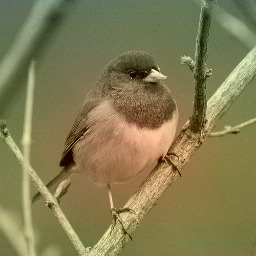}}  
\subfloat{\includegraphics[width=.08\textwidth]{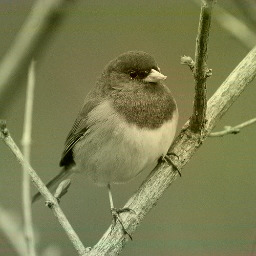}} \\ [-2ex]
\subfloat{\includegraphics[width=.08\textwidth]{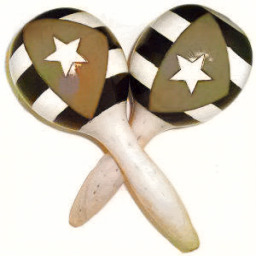}}  
\subfloat{\includegraphics[width=.08\textwidth]{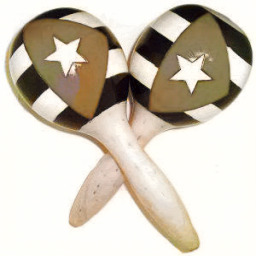}}  
\subfloat[cGAN]{\includegraphics[width=.08\textwidth]{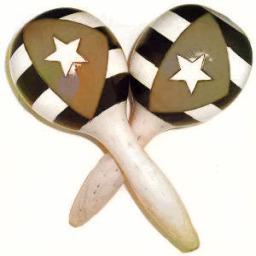}}  
\subfloat{\includegraphics[width=.08\textwidth]{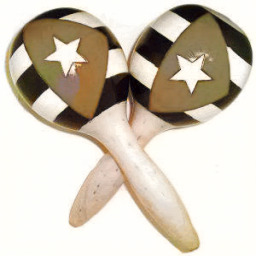}}  
\subfloat{\includegraphics[width=.08\textwidth]{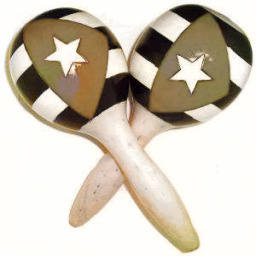}}  \hspace{8pt} 
\subfloat{\includegraphics[width=.08\textwidth]{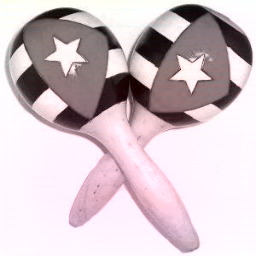}}  
\subfloat{\includegraphics[width=.08\textwidth]{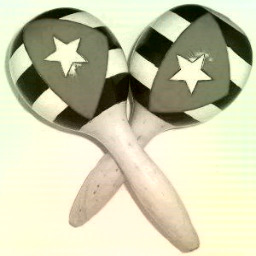}}  
\subfloat[CVAE]{\includegraphics[width=.08\textwidth]{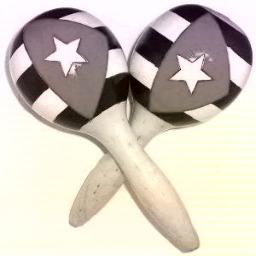}}  
\subfloat{\includegraphics[width=.08\textwidth]{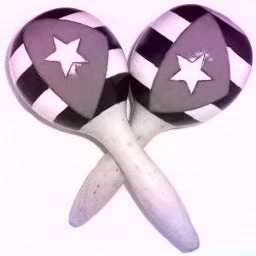}}  
\subfloat{\includegraphics[width=.08\textwidth]{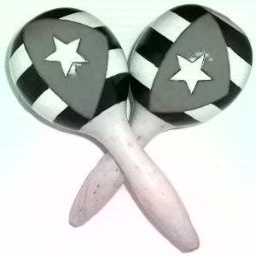}} \hspace{8pt}
\subfloat[GT]{\includegraphics[width=.08\textwidth]{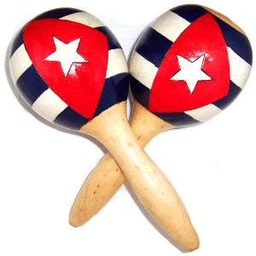}} \\[-2.5ex]
\subfloat{\includegraphics[width=.08\textwidth]{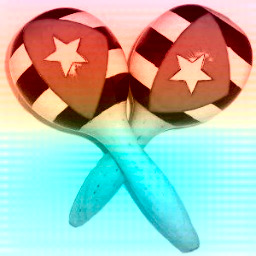}}  
\subfloat{\includegraphics[width=.08\textwidth]{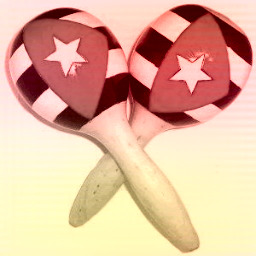}}  
\subfloat[Ours]{\includegraphics[width=.08\textwidth]{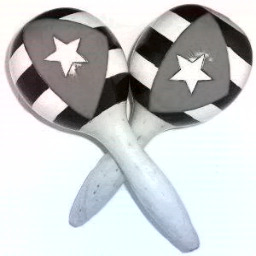}}  
\subfloat{\includegraphics[width=.08\textwidth]{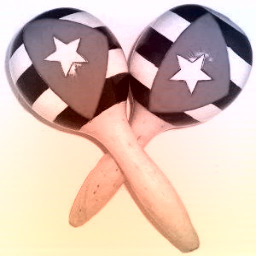}}  
\subfloat{\includegraphics[width=.08\textwidth]{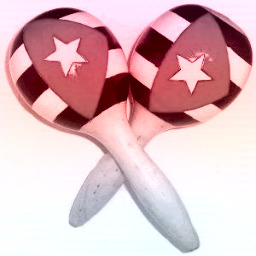}}  \hspace{8pt} 
\subfloat{\includegraphics[width=.08\textwidth]{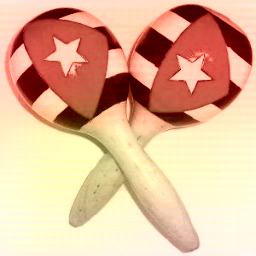}}  
\subfloat{\includegraphics[width=.08\textwidth]{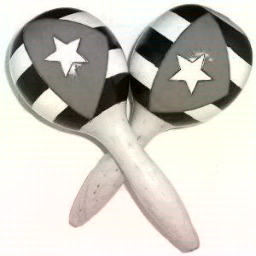}}  
\subfloat[Ours+Skip]{\includegraphics[width=.08\textwidth]{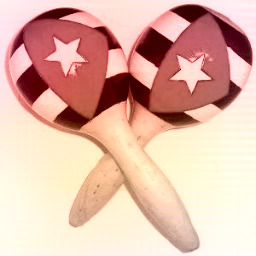}}  
\subfloat{\includegraphics[width=.08\textwidth]{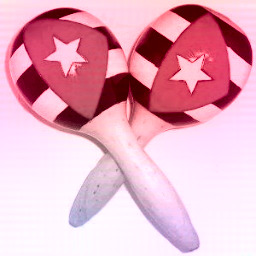}}  
\subfloat{\includegraphics[width=.08\textwidth]{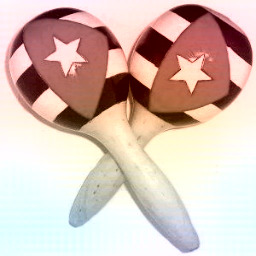}} \\ [-2ex]
\subfloat{\includegraphics[width=.08\textwidth]{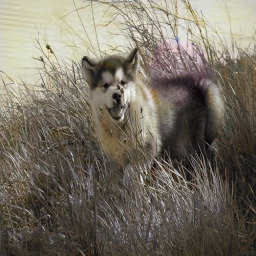}}  
\subfloat{\includegraphics[width=.08\textwidth]{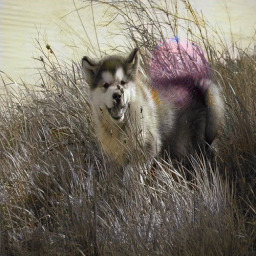}}  
\subfloat[cGAN]{\includegraphics[width=.08\textwidth]{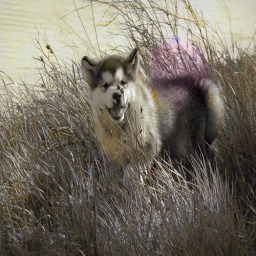}}  
\subfloat{\includegraphics[width=.08\textwidth]{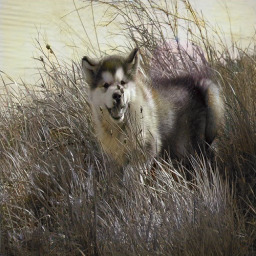}}  
\subfloat{\includegraphics[width=.08\textwidth]{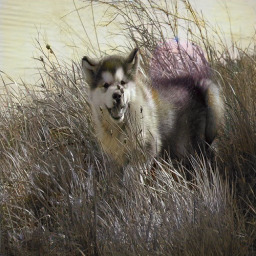}}  \hspace{8pt} 
\subfloat{\includegraphics[width=.08\textwidth]{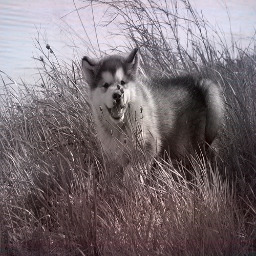}}  
\subfloat{\includegraphics[width=.08\textwidth]{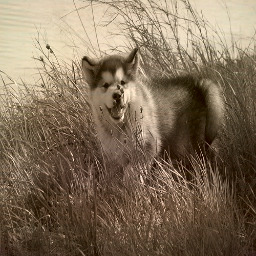}}  
\subfloat[CVAE]{\includegraphics[width=.08\textwidth]{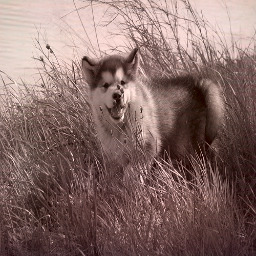}}  
\subfloat{\includegraphics[width=.08\textwidth]{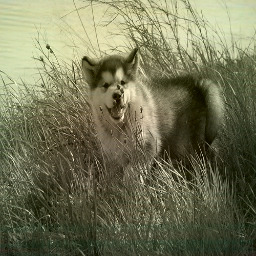}}  
\subfloat{\includegraphics[width=.08\textwidth]{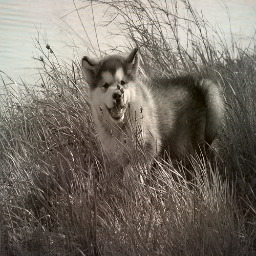}} \hspace{8pt}
\subfloat[GT]{\includegraphics[width=.08\textwidth]{imagenetval/gt/ILSVRC2012_val_00028070.jpg}} \\[-2.5ex]
\subfloat{\includegraphics[width=.08\textwidth]{imagenetval/cvpr_res/ILSVRC2012_val_00028070.JPEG/divcolor_000.jpg}}  
\subfloat{\includegraphics[width=.08\textwidth]{imagenetval/cvpr_res/ILSVRC2012_val_00028070.JPEG/divcolor_001.jpg}}  
\subfloat[Ours]{\includegraphics[width=.08\textwidth]{imagenetval/cvpr_res/ILSVRC2012_val_00028070.JPEG/divcolor_002.jpg}}  
\subfloat{\includegraphics[width=.08\textwidth]{imagenetval/cvpr_res/ILSVRC2012_val_00028070.JPEG/divcolor_003.jpg}}  
\subfloat{\includegraphics[width=.08\textwidth]{imagenetval/cvpr_res/ILSVRC2012_val_00028070.JPEG/divcolor_004.jpg}}  \hspace{8pt} 
\subfloat{\includegraphics[width=.08\textwidth]{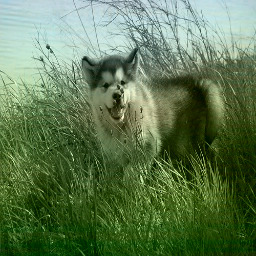}}  
\subfloat{\includegraphics[width=.08\textwidth]{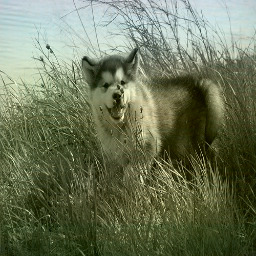}}  
\subfloat[Ours+Skip]{\includegraphics[width=.08\textwidth]{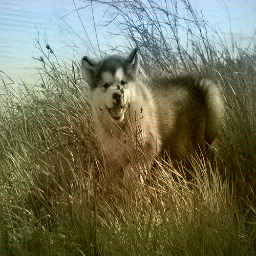}}  
\subfloat{\includegraphics[width=.08\textwidth]{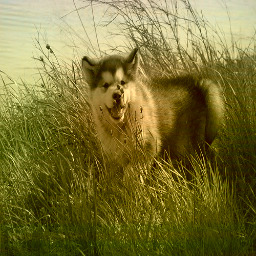}}  
\subfloat{\includegraphics[width=.08\textwidth]{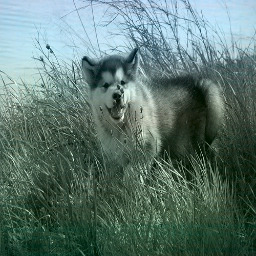}} \\ [-2ex]

\caption{For ImageNet-Val dataset, diverse colorizations from our methods are compared to the CVAE, cGAN and the ground-truth (GT).} 
\label{fig:res_imagenetval} 
\end{figure*}
\end{center}

\end{document}